# The Efficacy of Utility Functions for Multicriteria Hospital Case-Mix Planning

Robert L Burdett[1], Paul Corry[1], Prasad Yarlagadda[2], David Cook[3], Sean Birgan[3]

[1] School of Mathematical Sciences, Queensland University of Technology, Brisbane, Qld, Australia
[2] University of Southern Queensland, Toowoomba, Qld, Australia
[3] Princess Alexandra Hospital, 2 Ipswich Rd, Woolloongabba, Brisbane, Qld, Australia

**Abstract** – A new approach to perform hospital case-mix planning (CMP) is introduced in this article. Our multi-criteria approach utilises utility functions (UF) to articulate the preferences and standpoint of independent decision makers regarding outputs. The primary aim of this article is to test whether a utility functions method (UFM) based upon the scalarization of aforesaid UF is an appropriate quantitative technique to, i) distribute hospital resources to different operating units, and ii) provide a better capacity allocation and case mix. Our approach is motivated by the need to provide a method able to evaluate the trade-off between different stakeholders and objectives of hospitals. To the best of our knowledge, no such approach has been considered before in the literature. As we will later show, this idea addresses various technical limitations, weaknesses, and flaws in current CMP. The efficacy of the aforesaid approach is tested on a case study of a large tertiary hospital. Currently UF are not used by hospital managers, and real functions are unavailable, hence, 14 rational options are tested. Our exploratory analysis has provided important guidelines for the application of these UF. It indicates that these UF provide a valuable starting point for planners, managers, and executives of hospitals to impose their goals and aspirations. In conclusion, our approach may be better at identifying case mix that users want to treat and seems more capable of modelling the varying importance of different levels of output. Apart from finding desirable case mixes to consider, the approach can provide important insights via a sensitivity analysis of the parameters of each UF.

1.  Introduction

In this article case mix planning (CMP) in hospitals is focused upon. This strategic planning problem is important and sits at the top of a hierarchy of tactical and operational decision problems like operating room planning and patient scheduling (Leeftink and Hans 2018, Burdett et al 2018, Aringheiri et al 2022). The purpose of CMP is ultimately to support heath care managers improve resource management in hospitals (Hof et al. 2017, Ma and Demeulemeester 2013). The goal is to identify a patient caseload (a.k.a. cohort) to treat with a specific set of features deemed desirable or ideal (Andrews et al. 2022). This choice has significant economic consequences, and greatly affects the operation of a hospital, and the size of patient waiting lists. In the literature, the usual metrics for desirability are number of patients treated, the total revenue obtained, or the total costs incurred.

When performing CMP, it is first necessary to partition patients into a set of homogenous groups each with common characteristics (Landa et al., 2018). Each group may refer to a particular operating unit, medical or surgical speciality, a particular patient type, or patients with a particular condition and/or illness.  Depending upon the agenda of stakeholders, CMP can be modelled at different levels of detail. In most papers, hospital operations are modelled at a macroscopic level, and scheduling policies and other operational considerations are not included. This permits CMP to be performed over longer time horizons.  Some papers, however, consider greater levels of detail and provide microscopic planning and scheduling models. Those models, however, are time consuming, if not intractable to solve and rely upon a discretisation of the time horizon. If the time horizon is long and or the number of patients is large, then optimal solutions may not be guaranteed.

In past research, a variety of approaches have been applied to CMP, including goal programming (Blake and Carter 2002), mixed integer programming (Burdett et al. 2017), multicriteria optimisation (Malik et al. 2015, Burdett and Kozan 2016, Zhou et al. 2018, Chalgham et al. 2019), stochastic programming (Neyshabouri and Berg 2017, Freeman et al. 2018, McRae and Brunner 2019, Burdett et al. 2023c), and discrete event simulation (Oliveira et al. 2020). In addition, Leeftink and Hans (2018) have proposed a case mix classification scheme. Burdett et al. (2023b) have considered the needs of end users and developed a personal decision support tool. Table 1 below summarises the state of the art presently, and the main features that have been included to date.

**Table 1.** Focus of recent CMP research

| Article | Problem | ED | OR | WARD | ICU | STOCH | MC | GUI | REG | Objectives & Method |
|---|---|---|---|---|---|---|---|---|---|---|
| Ma et al. (2011) | CMS | ✗ | ✓ | ✓ | ✗ | ✗ | ✗ | ✗ | ✗ | Profit. MIP |
| Ma & Demeulemeester (2013) | CMP+ORS | ✗ | ✓ | ✓ | ✗ | ✗ | ✗ | ✗ | ✗ | Profit; Bed Shortage; MIP |
| Malik et.al. (2015) | ORP | ✗ | ✓ | ✗ | ✗ | ✗ | ✓ | ✗ | ✗ | Waiting List; Size; Costs; Meta H. |
| Jebali and Diabat (2015) | ORP | ✗ | ✓ | ✗ | ✓ | ✓ | ✗ | ✗ | ✗ | Costs; SAA |
| Burdett & Kozan (2016) | HCA | ✗ | ✓ | ✓ | ✓ | ✓ | ✗ | ✗ | ✗ | Output × 21; LP, ECM. |
| Yahia et al. (2016) | CMP | ✗ | ✓ | ✓ | ✓ | ✓ | ✗ | ✗ | ✗ | Output; SAA |
| Jebali and Diabat (2017) | ORP | ✗ | ✓ | ✗ | ✓ | ✓ | ✗ | ✗ | ✗ | Cost; SAA |
| Burdett et al. (2017) | HCA | ✓ | ✓ | ✓ | ✓ | ✗ | ✗ | ✗ | ✗ | Output; LP |
| Zhou et al. (2018) | HCA | ✗ | ✓ | ✓ | ✗ | ✓ | ✓ | ✗ | ✗ | Revenue; Equity; DES, MIP, ECM |
| Shafaei & Mozdgir (2018) | ORP | ✗ | ✓ | ✓ | ✓ | ✓ | ✗ | ✗ | ✗ | Value; LP; TOPSIS; |
| Freeman et al. (2018) | ORS | ✗ | ✓ | ✓ | ✓ | ✓ | ✗ | ✗ | ✗ | Payment; MIP |
| McRae et al. (2018) | CMP | ✗ | ✓ | ✓ | ✓ | ✗ | ✗ | ✗ | ✗ | Profit; NLP; UF |
| McRae & Brunner (2019) | CMP | ✗ | ✓ | ✓ | ✓ | ✓ | ✗ | ✗ | ✗ | Revenue; SAA |
| Saha & Rathore (2022) | CMP | ✗ | ✗ | ✗ | ✓ | ✓ | ✗ | ✗ | ✗ | Expected Cost; Heuristic |
| Burdett et al. (2023a) | HCA | ✗ | ✓ | ✓ | ✓ | ✗ | ✓ | ✗ | ✓ | Output, Unmet Demand, Outsourcing; MIP |
| Burdett et al. (2023b) | HCA | ✗ | ✓ | ✓ | ✓ | ✗ | ✗ | ✓ | ✗ | Output; HOPLITE, MIP |
| Burdett et al. (2023c) | HCA | ✗ | ✓ | ✓ | ✓ | ✓ | ✗ | ✗ | ✗ | Output; SAA, Meta H. |
| This article (2023) | HCA | ✗ | ✓ | ✓ | ✗ | ✓ | ✗ | ✗ | ✗ | Output; UF; |

**Key:** CMP: Case Mix Planning; CMS: Case Mix Scheduling; DES: Discrete Event Simulation; ECM: Epsilon Constraint Method; GUI: Graphical User Interface; HCA: Hospital Capacity Allocation; LP: Linear Programming; MIP: Mixed Integer Programming; ORS: Operating Room Scheduling; ORP: Operating Room Planning; REG: Regional; SAA: Sample Average Approximation; UF: Utility Function

Case mixes are an important concept in CMP. A case mix is the specific blend (a.k.a., mixture) of patients (i.e., to be imposed) within a cohort. Without intervention and planning, a hospital's case mix is dictated by the training, skills and interests of staff, the referral patterns of patients, the productivity of the hospital, and prevalence of disease within the catchment areas (Blake et al. 2002). The case mix is hierarchical. A further division of the patients within a particular group into sub-groups or sub-types is routine. As such, a case mix must be defined for each group to describe the relative number of patients of each of its subtypes.

The case mix is frequently an input to CMP (Burdett et al. 2017), defined upfront by decision makers and planners of hospitals. It is used as a mechanism to preference specific groups of patients over a planning horizon and a mechanism to regulate the competition for resources. There is, however, no universal definition, applicable to all situations. Each case mix definition produces a different caseload and a different profile of resource usage (Burdett et al. 2023b). In the literature, case mix is often viewed as the relative number of patients of each group or type that is treated. This means that for each group $g \in G$, there is a given proportion $\mu_g \in [0,1]$ such that $\sum_g \mu_g = 1$. Hence, we impose that the number of patients of type $g$ is governed by the equation $n_g = \mu_g N$ where $N = \sum_g n_g$. The main drawback of this approach is that if one group of patients is bottlenecked, then all the other groups of patients are too. Consequently, without altering the case mix designated by the user, it is not possible to use the latent capacity in the system to treat other groups of patients. In the language and terminology of multicriteria analysis, caseloads of this nature are called "dominated", as other caseloads exist (i.e., non-dominated) which permit the latent capacity to be used (Burdett et al. 2016).

Anecdotally we have observed that hospitals do not always view the case mix as described above. They often view case mix as a relative measure of the theatre time allocated to each surgical patient group or type. For instance, $n_g t_g = \mu_g T$ where $t_g$ is the average theatre time for group $g$ and $T$ is the total theatre time available. The case mix, however, can be defined relative to any hospital resource type.

CMP with an output focused objective is an inherently multicriteria decision problem with many-objectives. This is because each group of patients has conflicting interests, and shares resources (i.e., like operating theatres and in-patient beds) with other groups. Only on rare occasions is that not so. Without a formal mechanism such as a case mix, it is necessary to find an acceptable trade-off another way, for instance by obtaining and analysing the Pareto frontier of alternatively optimal (i.e., non-dominated) solutions. Each non-dominated solution describes a completely different trade-off and divides the time availability (a.k.a., capacity) of each hospital resource, amongst the different groups of patients in a unique way. The drawback of such an approach is evident. When there are many patient types, the resulting multicriteria decision problem has a high number of dimensions, i.e., one for group of patients. As shown in Burdett and Kozan (2016), the number of Pareto optimal solutions that can be identified is excessive, and techniques like the epsilon-constraint method are inadequate.

***Research Agenda.*** There are many hospital stakeholders and hospital objectives, and it is important to provide methods to evaluate the trade-off between them. An important aspiration of most hospitals is to treat as many patients as possible of each type, within a given time horizon. However, between upper and lower base-levels of achievement, outputs are selectable and negotiable. To facilitate the best CMP, we ought to define a utility (a.k.a., achievement) function, that more clearly articulates our preferences and standpoint and those of decision makers (DM), regarding outputs. To the best of our knowledge, no such approach has been considered before in the CMP literature. As we will later show, this idea addresses various technical limitations, weaknesses, and flaws in current CMP. The following research questions are posed:

  i.   Conceptually how useful are utility functions for CMP activities?
  ii.  Can the use of utility functions negate the need to apply traditional multicriteria analysis and optimization techniques?
  iii. Are utility functions conceptually a better/worse approach than designating a case mix of some form and imposing case mix constraints?
  iv.  How are utility functions defined, altered, and renegotiated in an iterative CMP approach? Is there a more rigorous approach or set of guidelines to do so?
  v.   How do the results of CMP change when different utility functions are applied? What is the exact difference between the UF types?

The rest of the paper is organized as follows. In Section 2 the current state of the art is examined, and important background methodological information is provided. In Section 3 the details of the quantitative framework are provided. In Section 4, a case study of real-world size is presented. Last, the conclusions, managerial insights and future research directions are detailed. This article has numerous acronyms, and a summary can be found in Appendix A. In Appendix B and C technical details are provided. Important results are summarised in Appendix D.

**2. Methodological Background**

As a foundation for later developments a review of utility functions and salient multicriteria optimization techniques is provided in this section.

**Multicriteria Analysis and Optimization (MCO)**. Multicriteria analysis is an iterative process supporting the user in the exploration of a Pareto set consisting of non-dominated solutions. It aims at finding subsets of solutions with desired properties (Makowski 2009). MCO is the solution of a mathematical programming model with two or more objectives. Numerous methods have been developed for MCO and there are two main strategies. The first strategy involves the application of multi-objective programming methods to first find efficient solutions for the DM to choose from. This is known as a Pareto frontier (PF). In the second strategy, an auxiliary parametric single-objective model is posed, whose solution provides a single Pareto-optimal point (Granat and Makowski, 2000). In the second strategy utility functions are predominantly applied. Eliciting preference information from the DM is first necessary to construct a utility function which is subsequently optimized. The preferences of DMs may then change as they learn more of the decision situation (Stewart (1996)). Ehrgott et al. (2009) is noteworthy for comparing both strategies for portfolio optimization with multiple objectives. They generated efficient solutions upfront for the investor and applied utility functions to optimize a single objective mathematical programming model.

**Utility Functions.** A Utility Function (UF) is a relative measure of the desirability (i.e., global utility) of different alternatives. They are often used to measure preferences concerning goods and services. It has been said that every decision maker (DM) tries to optimize, consciously or unconsciously, a utility or payoff function aggregating all their points of view (Wierzbicki 1977).

Utility Function Methods (UFM) and Value Function Methods are techniques that apply utility functions to MCO problems. Assuming $m$ objectives and $n$ decision variables, the objective is to maximize $U(f(x))$, such that $x \in X$, i.e., to choose $x_{opt} = arg \max_{x \in X} U(f(x))$. Here, $x$ is a decision vector, $X$ is the decision space, $U$ is a utility function that maps $\mathbb{R}^m \to \mathbb{R}^1$, and $f$ is a function to evaluate the different objectives, that maps $\mathbb{R}^n \to \mathbb{R}^m$. It is worth noting that $U(f(x_a)) > U(f(x_b))$ implies solution $a$ is preferred to solution $b$. The UF must be strongly decreasing; this means the preference must increase if one objective is decreased, and all others are kept the same. Both additive and multiplicative models have been posed in the literature as shown in equation (1). In that equation $f_i$ computes the ith objective value and $u_i$ is the utility function for objective $i$ that maps $z_i$ to a particular achievement level. There are various assumptions related to the application of these, the foremost being mutual preferential independence (Keeney, 1971).

$$U(f(x)) = \prod_i u_i(z_i)) \text{ or } U(f(x)) = \sum_i u_i(z_i) \text{ where } z_i = f_i(x) \tag{1}$$

To obtain utility functions $u_i$ for real world MCO problems, numerous approaches have been proposed. In recent years, interactive learning procedures have become trendy. Dewancker et al. (2016) proposed a generative "multiplicative" model and a machine learning approach. Shavarani et al. (2021) applied an interactive multi-objective evolutionary algorithm and use a proven sigmoidal UF. Interactive methods are designed to explore interesting parts of the Pareto Frontier. They comment that simple and/or unrealistic utility functions are most often applied in published articles to simulate the behaviour of real decision makers (Shavarani et al. 2021). Torkjazi and Fazlollahtabar (2015) considered the application of multiple utility functions per objective in a MCO problem. They applied fuzzy probabilistic programming techniques. Multiple utility functions are deemed necessary because there is imprecision, uncertainty, and ambiguity in those functions. It is also noted that utility functions may be defined for each objective based on different situations and different environments. Longaray et al (2018) proposed a multicriteria decision analysis to evaluate the performance of activities in the internal logistics process of the supply chain of a teaching hospital. They used the categorical based evaluation technique (MACBETH) to generate value functions. The MACBETH method aggregates performance values in different criteria using an additive value function model.

**Goal Attainment (GAM) and Goal Programming Methods (GPM).** These methods have been applied to multicriteria decision problems for numerous years (Gur and Eren, 2018). Some new versions have been created recently, however. In Gur and Eren (2018) the number of goals that are reached is maximized. This is called extended goal programming (EGP). In Hezam et al. (2022) a goal programming approach based upon fuzzy logic theory is proposed for evaluating the resources of health organisations. In their healthcare planning problem, they included staffing levels, medical supplies and drugs, staff rostering, and budgets. They applied their model to an oncology centre.

Although described as clear and appealing, these methods *are criticized* by MCO specialists for their non-compliance with the Pareto optimality principle (Ogryczak and Lahoda, 1992). In the literature the models presented in (2)-(5) are most prevalent. They have been previously posed in the literature and can be found summarised in sources like Ogryczak and Lahoda (1992) and Stewart (2005). Multiplicative versions can also be posed by changing the summations to products.

$$\text{Minimize } \left\{ \epsilon_1 \left( \max_i (w_i \delta_i) \right) + \epsilon_2 \sum_i (w_i \delta_i) \right\} \text{ s.t. } z_i = d_i - \delta_i \text{ or } z_i = d_i + \delta_i \text{ and } \delta_i \geq 0 \quad (2)$$

$$\text{Minimize } \sum_{i \in I} w_i |z_i - d_i| \text{ or } \sum_{i \in I} w_i \left| \frac{z_i - d_i}{d_i} \right| \quad (3a)$$

$$\text{Minimize } \sum_i (w_i^+ \delta_i^+ + w_i^- \delta_i^-) \text{ s.t. } z_i - d_i = \delta_i^+ - \delta_i^-, \; \delta_i^+ \delta_i^- = 0 \text{ and } \delta_i^+, \delta_i^- \geq 0 \; (GPM) \quad (3b)$$

$$\text{Minimize } \left\{ \max_i (w_i^+ \delta_i^+ + w_i^- \delta_i^-) \right\} \text{ or } \left\{ \max_i (w_i^+ \delta_i^+) + \max_i (w_i^- \delta_i^-) \right\} \quad (3c)$$

$$\text{Minimize } \delta \text{ s.t. } z_i \geq d_i - w_i \delta, \; z_i \leq d_i + w_i \delta \text{ and } \delta \geq 0 \quad (GAM) \quad (4a)$$

$$\text{Minimize } \epsilon^+ \delta^+ + \epsilon^- \delta^- \text{ s.t. } z_i \leq d_i + w_i \delta^+, \; z_i \geq d_i - w_i \delta^- \text{ and } \delta^+, \delta^- \geq 0 \quad (4b)$$

$$\text{Minimize } \sum_i \max(0, z_i - d_i - w_i \delta) \; \text{ for a given } \delta \quad (4c)$$

$$\text{Minimize } \sum_i \max(0, d_i - z_i) \text{ or } \sum_i \max(0, z_i - d_i) \quad (5)$$

These models are solved subject to other problem specific decision variables and technical constraints. For objective $i$, $d_i$ is the goal (a.k.a., aspiration), the under-achievement is defined by $\delta_i^-$ and the over-achievement by $\delta_i^+$. The term $\delta$ is used generally for either type of deviation. In (2), under and over-achievements respectively are penalized, but not both. Depending upon how $\epsilon_1$ and $\epsilon_2$ are defined, the aggregate deviation and the extent of the worst over or under-achievement (i.e., the Chebyshev utility function) can be minimized. We could choose $\epsilon_1 \leq 1$ and $\epsilon_2 \leq 1$, such that $\epsilon_1 + \epsilon_2 = 1$. We could also set them independently. The same could be said of $\epsilon^+$ and $\epsilon^-$ in (4b). In (3a), any deviation (scaled or unscaled) is deemed undesirable. This is recognisable as the 1-norm. Option (3b) is a variant of (3a) that weights over and under-achievements differently. This is the traditional *Goal Programming Method* (GPM). In (3c), the maximum over and under-achievement are minimized. Although not shown, (3a) or (3b) could be aggregated with (3c). In (4a) and (4b), terms $w_i \delta$, $w_i \delta^+$ and $w_i \delta^-$ introduce an element of slackness into the problem, so that goals do not need to be rigidly met. In (4a), over or under-utilizations respectively are minimized. This is the traditional *Goal Attainment Method* (GAM). In (4b), both over and under-utilization are considered, but regarded with potentially different importance. In (4c), aggregate over-achievement is minimized. All over-achievements are, however, permitted, in contrast to (4b) which explicitly sets hard limits. Some over-achievements are permitted without penalty, and do not contribute to the score. This is governed by parameter $\delta$. Objective (5) is a variant of (4c), from Benson (1978).

*The biggest issue* with most of these methods is that deviations are penalised in a static way, when really the penalty should increase as the deviation becomes bigger. In other words, slight under or over-achievements are inconsequential, but larger ones are not. These methods meet goals as best possible, but do not consider if any can be exceeded. Hence, solutions are not necessarily Pareto optimal.

**Aspiration Reservation Method (ARM).** The ARM is an approach for multi-criteria analysis of decision problems. It has been well exemplified and sponsored in the literature for instance by Wierzbicki (1977), Granat and Makowsi (2000, 2006), Makowski (2009). It is an arguably a better approach than the GAM and GPM. Simple UF (i.e., with one or two piecewise linear segments) are the backbone of the ARM and are used to articulate more clearly the preferences of decision makers. In the ARM, each objective is given a criterion (a.k.a., component) achievement function (CAF) and multiple objectives are aggregated into one objective, using an achievement scalarizing function (ASF), that maps $R^n \to R^1$. Important concepts are the aspiration $z_i^a$ and reservation (a.k.a., reference) points $z_i^r$. The former is a solution composed of the desired values for the corresponding criterion. The latter is a solution composed of acceptable values for the corresponding criterions. The traditional ASF is maximized and defined as follows, $ASF = \min_i\{u_i(z_i,.)\} + \epsilon \sum_i u_i(z_i,.)$, where $\epsilon$ is a small value, $z_i$ is the value of the ith objective, and $u_i(z_i)$ is a UF / CAF. For the ARM, it is necessary to define functions $u_i(z_i, z_i^a, z_i^r)$ for all $i \in I$. According to Ogryczak and Lahoda (1992), solutions that satisfy all aspiration levels are preferable to outcomes that do not in the ARM method. Given strict upper and lower limits, it is beneficial to normalize the achievement as follows: $u_i(z_i, \bar{z}_i, \underline{z}_i) = (z_i - \underline{z}_i)/(\bar{z}_i - \underline{z}_i)$.

As described, the ARM *is quite basic*. A more general version can be implemented, without any notion of aspiration and reservation points. General utility functions with more piecewise linear segments or non-linearities may be used. In the next section, that approach is taken.

## 3. Quantitative Framework for CMP

In this section the case-mix planning model is first introduced before a utility function method is proposed.

### 3.1. The CMP Model

In this article we choose to consider a high-level strategic CMP problem. It is described by the optimization model shown in (6)-(12). The purpose of this model is to identify the number of patients of each type (a.k.a., group) and subtype (a.k.a., sub-group) to treat over time, denoted respectively $n_g^1$ and $n_{g,p}^2$, given the current hospital configuration and some basic patient resourcing requirements. These variables are rates of output and do not refer to discrete patients. There is an inherent hierarchy between $n_g^1$, and $n_{g,p}^2$, namely $n_g^1 = \sum_{p \in P_g} n_{g,p}^2$, and $P_g$ is the set of subtypes within group $g$.

The resourcing requirements for each patient subtype $(g, p)$ is described by a resourcing profile (a.k.a., patient care pathway). This resourcing profile is just a list of activities. For the purposes of this paper, and for a high-level capacity modelling perspective, the sequence of events, is irrelevant. The output of the hospital, denoted $\mathbb{N}$, is restricted by the resources present, their time availability, and their purpose. As such, it is necessary to identify a resource allocation, describing which resources will be used to treat each patient. The resource allocation is denoted by $\beta_{a,r}$. This decision variable describes how many patients with activity $a$ are treated by resource $r$.

The model has various bookkeeping constraints. Constraint (7) defines the inherent relationship between the number of patients $n_{g,p}^2$ and the resource allocation. Resource usage is restricted by the time availability of the resource as shown in constraint (8). Designated case mix are enforced by (9) and (10) if needed. The remainder, namely (11) and (12) enforce positivity.

Maximize $\mathbb{N} = \sum_{g \in G} n_g^1 = \sum_{g \in G} \sum_{p \in P_g} n_{g,p}^2$ (6)

Subject To:

$n_{g,p}^2 = \sum_{r \in R_a} \beta_{a,r} \quad \forall g \in G, \forall p \in P_g, \forall a \in A_{g,p}$ (7)

$\sum_{a \in A_r} \beta_{a,r} t_a \leq T_r \quad \forall r \in R \text{ where } T_r = h_r \times \mathbb{T}$ (8)

$n_g^1 \geq \mu_g^1 \sum_g n_g^1 \quad \forall g \in G$ (9)

$n_{g,p}^2 \geq \mu_{g,p}^2 n_g^1 \quad \forall g \in G, \forall p \in P_g$ (10)

$n_g^1, n_{g,p}^2 \geq 0 \quad \forall g \in G, \forall p \in P_g$ (11)

$\beta_{a,r} \geq 0 \quad \forall a \in A, \forall r \in R_a \text{ and } \beta_{a,r} = 0 \quad \forall a \in A, \forall r \in R \backslash R_a$ (12)

To fully understand this model, it is also necessary to point out the following:
  i. $\mathbb{T}$ is the period of planning, i.e., the number of weeks considered.
  ii. $R$ is the set of resources. We only consider hospital facilities such as operating theatres, wards, and intensive care in this article. Auxiliary resources like staffing could also be integrated.
  iii. $A_{g,p}$ is the set of activities for patient subtype $(g, p)$. Hence, $A_g = \bigcup_{p \in P_g} A_{g,p}$. In addition, $A$ is the complete set of activities. As such: $A = \bigcup_{g \in G} A_g$.

- iv. $R_a \subset R$ is the resourcing profile for activity $a$, i.e., the set of resources that can be used. This set is defined relative to the type of activity being performed.
- v. $t_a$ is the time to perform activity $a$.
- vi. $h_r$ is the time availability weekly of resource $r$. If the resource is a facility like a ward or intensive care unit, then this number must be multiplied by the number of beds present.
- vii. The patient type mix and sub mix are denoted $\mu_g^1$ and $\mu_{g,p}^2$ respectively where $\sum_g \mu_g^1 = 1$ and $\sum_{p \in P_g} \mu_{g,p}^2 = 1$.
- viii. Upper bounds designated by $\bar{n}_g^1$ are important to compute. The upper bound is determined from the CMP model, assuming the following single patient case mix, $\mu_g^1 = 1; \mu_{g'}^1 = 0 \ \forall g' \in G \setminus \{g\}$.

## 3.2. Solving the Multicriteria CMP Problem

The multicriteria CMP problem considers the maximization of each patient type simultaneously. In the CMP model, case mix constraint (9) is omitted, and objective function (6) is conceptually replaced with the following: Maximize $\{n_1^1, n_2^1, \ldots, n_{|G|}^1\}$. In this section goal programming methods and utility functions are revisited as a means of navigating the Pareto frontier of the multicriteria CMP problem. These methods permit conversion to a single objective. They can be used to minimize the level of over or under-achievement from imposed goals denoted $\hat{n}_g^1$ and to maximize total treatments given by $\sum_{g \in G} n_g^1$. The relevant details of each are now discussed:

**3.2.1. Goal Attainment.** The goal attainment model for CMP is summarised by equation (13)-(16). The domain of the goals, $\hat{n}_g^1$ is $[0, \bar{n}_g^1]$. In a multi-criterion CMP setting, it is worth setting $\hat{n}_g^1 = \bar{n}_g^1$.

Minimize $\delta$ (13)
Subject to:
$\delta \geq 0$ (14)
$n_g^1 \leq \hat{n}_g^1 + w_g \delta \ \forall g \in G$ (i.e., $(n_g^1 - \hat{n}_g^1)/w_g \leq \delta$) (15)
$n_g^1 \geq \hat{n}_g^1 - w_g \delta \ \forall g \in G$ (i.e., $(\hat{n}_g^1 - n_g^1)/w_g \leq \delta$) (16)

The permitted deviation is governed by $\delta$ and the group specific priority $w_g$. Equation (15) restricts over-achievements if $w_g > 0$, and equation (16) restricts under-achievements. The sign of parameter $w_g$ is important. In (15), if $w_g < 0 \ \forall g \in G$, then no goal can be reached. In (16), if $w_g < 0 \ \forall g \in G$, then every goal must be exceeded. It's worth noting that if all goals are achievable, then $\delta = 0$.

Practically, we could set all $w_g = 1$, and this would then restrict the output of all groups in the same way. If any $w_g = 0$, then achievement is a hard constraint. Hence, smaller values imply less freedom to deviate from goals, and larger values permit the opposite. If $w_g$ are different, then different levels of over-achievement may be permitted for some groups of patients. In other words, some groups of patients would be prioritized. As reported at MathWorks (2022), if $w_g = \hat{n}_g^1$, then we restrict the relative achievement in the following way:

$$\frac{n_g^1}{\hat{n}_g^1} \leq (1 + \delta) \text{ and } \frac{n_g^1}{\hat{n}_g^1} \geq (1 - \delta) \quad (17)$$

**3.2.2. Goal Programming.** The goal programming method summarised by (18) is like the GAM, but a single parameter $\delta$ is not used to restrict the output of all groups.

Minimize $\sum_{g \in G}(w_g^+ \delta_g^+ + w_g^- \delta_g^-)$
Subject to: $n_g^1 = \hat{n}_g^1 + \delta_g^+ - \delta_g^-$, $\delta_g^+ \delta_g^- = 0$ and $\delta_g^+, \delta_g^- \geq 0 \ \forall g \in G$ (18)

This is essentially a weighted sum approach. The results are highly dependent upon parameters $w_g^+$ and $w_g^-$. As such, priority will be given to some groups, and not others. It is also reasonable to optimize the relative under and over-achievement, $\sum_{g \in G}(w_g^+ \acute{\delta}_g^+ + w_g^- \acute{\delta}_g^-)$ where $\acute{\delta}_g^+ = \delta_g^+/\hat{n}_g^1$ and $\acute{\delta}_g^- = \delta_g^-/\hat{n}_g^1$. This normalisation scales the differences and makes comparison between different groups more even-handed. To handle the non-linear constraint $\delta_g^+ \delta_g^- = 0$, it is necessary to incorporate a binary decision to force $\delta_g^+ = 0$ or $\delta_g^- = 0$. Let us define $\lambda_g = 1$ if $\delta_g^- = 0$, and zero if $\delta_g^+ = 0$. The following constraints are then required:

$$\delta_g^+ \leq \lambda_g(\bar{n}_g^1 - \hat{n}_g^1) \text{ and } \delta_g^- \leq (1-\lambda_g)\hat{n}_g^1 \ \forall g \in G \tag{19}$$
$$\lambda_g \in \{0,1\} \ \forall g \in G \tag{20}$$

**3.2.3. Utility Function Method (UFM).** An approach using utility functions, inspired by the ARM can be applied. The main assumptions are as follows:

i. We consider a single attribute, namely patient type, and define the attribute level as the number of patients treated.
ii. Each stakeholder represents a particular specialty and describes their level of satisfaction regarding different levels of output. They do not comment about the output of other specialties. Stakeholders who represent more than one specialty, including those that represent all specialties, are not considered.
iii. The preference of different levels of patient type $g$ do not depend on the levels of any other type $g'$. In other words, there is utility independence.

Given the above details, the application of objective function (21) with constraint (7)-(12) is appropriate:

$$\text{Maximize } ASF = \epsilon_1 \min_{g \in G}\{w_g u_g\} + \epsilon_2 \sum_{g \in G} w_g u_g \text{ where } u_g = \text{PLF}(n_g^1, b_g, \nabla_g) \tag{21}$$

In (21), $b_g$ are the breakpoints of the $g$th utility function, and $\nabla_g$ are the gradients of the line segments. We may, however, define the utility functions directly. For instance, the simplest options are $u_g = n_g^1$, $u_g = n_g^1 - \hat{n}_g^1$ and $u_g = (n_g^1 - \hat{n}_g^1)/\hat{n}_g^1$. The first option defines achievement as the weighted raw output, the second as the weighted difference from the aspiration, or thirdly as the weighted relative difference. A special case of (21) is when $\epsilon_2 = 0$. It maximizes the utility of the worst performing group.

**3.2.4. Generating Pareto Optimal Solutions.** Goal programming and goal attainment methods have known limitations in the context of multicriteria optimization. These methods minimize the over and under-achievement from the specified goals, and there is no incentive to do better, if the goals are achievable (i.e., the goals describe a dominated solution). In other words, it is possible to find non-dominated solutions. Figure 1 demonstrates for a basic two group scenario, where $n_1 \leq \bar{n}_1$ and $n_2 \leq \bar{n}_2$, the possibility of setting goals (i.e., A, B) above and below the implied Pareto frontier, that demarcates in the objective space, the boundary between feasibility and infeasibility. Goal B is not achievable, so the GAM and GPM must return the "nearest" feasible solution. That depends upon the weights used in the objective. That solution must be Pareto optimal, because any solution below the frontier would be non-optimal. Goal A is a dominated solution to the problem and would be the reported solution if the GAM or GPM were applied. To find a better solution, either of the models described at (23) and (24) could be applied in a "follow-up stage".

$$\text{Maximize } n_{g^*}^1 \text{ s.t. } n_g^1 \geq \hat{n}_g^1 \ \forall g \in G\setminus\{g^*\} \tag{23}$$
$$\text{Maximize } \Psi^+ = \sum_{g \in G}(w_g^+ \delta_g^+) \text{ s.t. } n_g^1 \geq \hat{n}_g^1 \ \forall g \in G \text{ or } \delta_g^- = 0 \ \forall g \in G \tag{24}$$

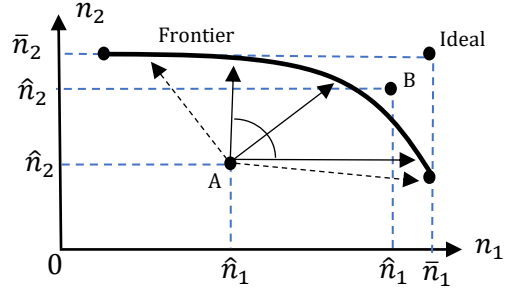

**Figure 1.** Goal setting above and below the Pareto frontier

In Figure 1, the solid black arcs show solutions that could be obtained. The model described at (23) permits a user to preference one of the patient groups, denoted $g^*$. Other patient groups will be kept at their respective goal level (i.e., $n_g^1 = \hat{n}_g^1$) if they share resources with patient group $g^*$. Otherwise, the goals can be exceeded for those patient types as well. In contrast, the model described at (24) does not explicitly preference any patient group. Instead, it seeks to optimise the overall improvement. Another approach is to solve a variant model that minimizes under-achievement and maximizes over-achievement. For instance:

$$\text{Maximize } \epsilon^+ \Psi^+ - \epsilon^- \Psi^- \tag{25}$$
Subject to: Constraint (18), (19), (20)

Where $\Psi^- = \sum_{g \in G}(w_g^- \delta_g^-)$ or $\Psi^- = \max_g\{w_g^- \delta_g^-\}$ and $\Psi^+ = \sum_{g \in G}(w_g^+ \delta_g^+)$ or $\Psi^+ = \min_g\{w_g^+ \delta_g^+\}$. In contrast to (23) and (24), this model may permit under-achievements, so that other greater over-achievements are realised. To avoid that happening, we can set $\epsilon^-$ to be a large value and $\epsilon^+ \cong 1$. In Figure 1, the dotted black arcs show possible solutions that could be obtained. In theory this model could supersede the others and be used for both stages described previously.

### 3.3. Utility Functions for CMP

To apply the UFM, it is necessary to define UF for each group of patients $g \in G$, or other category of interest. In each UF, a level of achievement must be defined for each conceivable number of patients treated. The UF may be defined as specific mathematical functions, otherwise they must be elicited from end users. Elicited UF may describe end users' subjective view of achievement relative to output. The achievement can be viewed in numerous ways. It can represent levels of satisfaction or dissatisfaction (unit = %), profit or loss (unit = $), achievement or non-achievement (unit = real value). The achievement function can be based upon quantitative data or qualitative.

The simplest utility function that may be used for CMP describe increased achievement and merit for increased treatments. A minimum requirement and aspiration can also be defined and incorporated. Any demand or target can be viewed as an aspiration; however, they may also be regarded as strict requirements. The principal options are shown in Figure 2. The x-axis is the output, and the y-axis is the metric of achievement. Most of the UF in Figure 2 have only two or three linear segments. Those with one may be characterized as simple functions, and those with more called compound functions. It is worth noting that UF1 is a special case of UF2, UF3 and UF4. Similarly, UF2 is a special case of UF4, and UF3 is a special case of UF4. The functions UF4 and UF7 also have a similar shape.

Non-linear and non-monotonic variants (i.e., like UF6) are also shown. Convex and concave variants are specifically dotted. For modelling purposes, the nonlinear variants need to be broken up into an arbitrary number of sub segments. Fig 2(e)-(g) are more sophisticated variants of (a)-(d) that impose negative achievement for not meeting a minimum expectation, and reduced achievement for exceeding aspirations too greatly. Fig 2(h) explicitly shows the well-known s-shaped function, positioned around a reference point. In the literature gains are often perceived as concave, and losses convex. The perception of what constitutes gain and loss, however, is subjective. Last, Fig 2(i)-(k) show discontinuous utility functions with tiers. Fig 2(k) demonstrates how a strict requirement may be described.

Regarding these utility functions, two concepts are worth noting. The value of output above which achievement is first acknowledged, is called the point of indifference (PTOI) (a.k.a., the intercept). All values of output below this are deemed zero or negative. Another important value is the minimum output above which no further achievement is regarded. This has been called an *aspiration point* (ASPT) in past research. There is only a single aspiration for any given group. In UF4 and UF7, it is worth noting the point where utility is halfway. It is referred to as a reference point.

**Quantification.** The UF shown in Figure 2 can be described by explicit mathematical functions or as piecewise linear functions with a distinct number of linear segments. The details of appropriate options are summarised in Appendix B and C. In Appendix C, some non-linear variants are described, but this list does not include all possibilities. For the non-linear functions shown, the parameter $\alpha$ is used to alter the convexity of the curve. The ordered set of breakpoints for each group is defined as $b_g$ and the slopes as $\nabla_g$. Formally, the point of indifference is denoted $n_g^I$ and the aspiration as $n_g^A$. All non-linear UF need to be converted into piecewise-linear equivalents for the CMP model to be solved using commercial solvers like IBM ILOG CPLEX. The number of points in each UF can vary but a minimum of two is required.

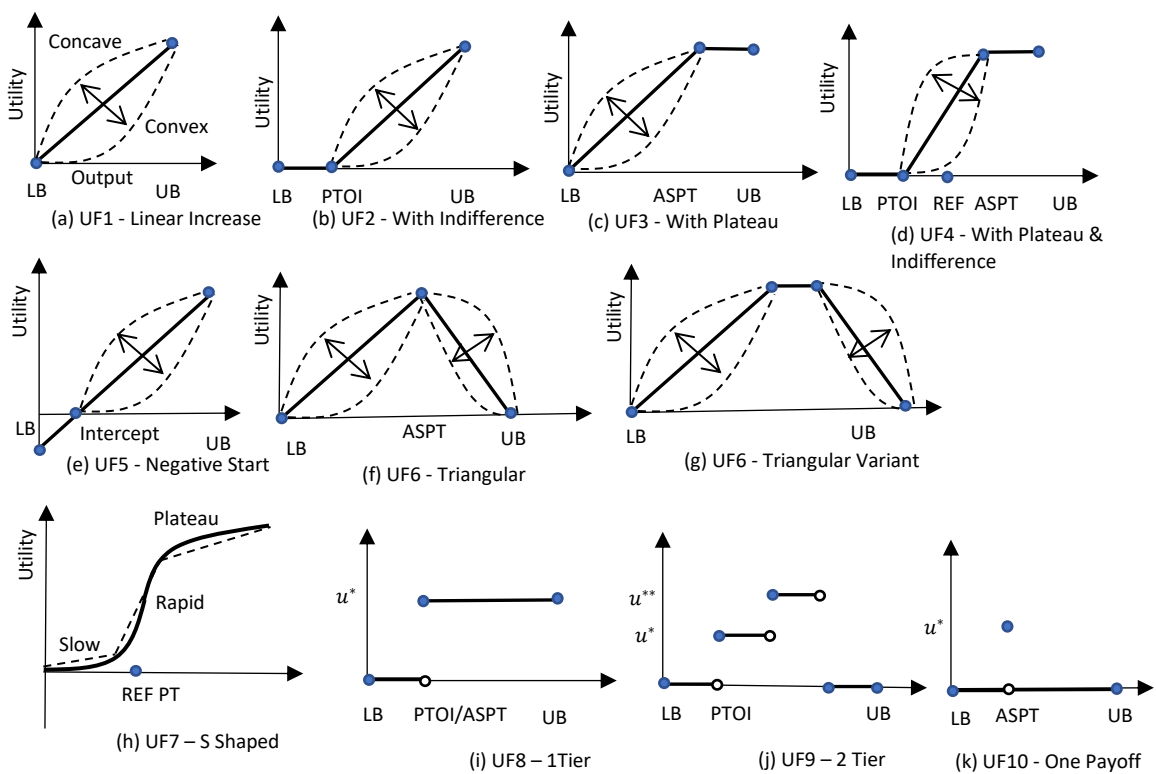

**Figure 2.** Piecewise linear and non-linear utility functions of practical relevance

To implement the tiered UF or other discontinuous piecewise linear functions, there are a few options. In IBM ILOG CPLEX, discontinuous piecewise linear functions can be accommodated by duplicating breakpoints and defining "jumps". Generally, any discontinuity occurring with points $(x, y)$ and $(x, y')$ is represented with two breakpoints $\{x, x\}$ and one slope, $\{y' - y\}$. In the absence of that functionality, these types of UF can be handled by adding an extra breakpoint before or after the discontinuity (i.e., $\{x - eps, x\}$ or $\{x, x + eps\}$), to create a steep sloped segment in place of the vertical one. Another approach is to incorporate additional auxiliary binary variables and constraints.

Let us define $\delta_{g,i}$ as one if the $i$th interval of the UF is selected, such that $\sum_{i\in\{1..I\}} \delta_{g,i} = 1$. To select the correct value, it is necessary to impose the following constraints:

$$l_{g,i} + M(\delta_{g,i} - 1) \leq n_g^1 \leq r_{g,i} + M(1 - \delta_{g,i}) \tag{27}$$
$$\text{PLF}_{g,i}(n_g^1) + M(\delta_{g,i} - 1) \leq u_g \leq \text{PLF}_{g,i}(n_g^1) + M(1 - \delta_{g,i}) \tag{28}$$

Above, $l_{g,i}, r_{g,i}$ are the input left and right boundary respectively for segment $i$ and $\text{PLF}_{g,i}(n_g^1) = m_{g,i} n_g^1 + c_{g,i}$.

**Pitfalls.** Utility functions may be defined for a given group without any understanding of the maximum number of patients that may be treated of that type in the hospital given full access to resources. This has implications for the proposed approach and for end users. For instance, if the chosen upper bound is less than the actual capability of the hospital, then an unfair limitation is imposed, one that may allow other groups of patients to capitalize. It is also possible to define aspirations that are above the capacity of the hospital. This may affect the analysis, because the upper bound if selected as the output, will be given a lower achievement level (i.e., see Fig 3). In fact, the upper bound is the aspiration point in these circumstances.

Given these two issues, a pre-analysis should be applied upfront to identify all upper bounds. Otherwise, a sufficiently large value should be identified. Also, we should enforce any aspiration point to be less than or equal to the upper bound.

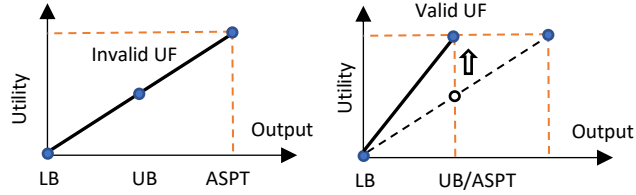

**Figure 3.** An invalid UF with an incorrectly defined aspiration value

The utility function in Fig 2b should be used with care. Conceptually, the point of indifference should be a small value. However, if the point of indifference is designated too high (i.e., by accident or design), for each group, then the solution $n_g^1 = n_g^I \; \forall g \in G$ may be infeasible. If objective function (21) with $\epsilon = 0$ is selected, then the optimal solution may be a zeroed (i.e., null) solution. This is unhelpful given that there is capacity to achieve more output. This, however, is logical, as in the range $[0, n_g^I]$ all solutions have the same utility; none. So, there is no point in choosing a higher output below the point of indifference. In conclusion, it would be better to have two sloping segments, within the utility function. The first segment would have a lesser slope, than the second. In other words, a piecewise linear version of the non-linear option shown in Fig 2c is suggested.

**Final Remarks**. An alternative viewpoint may be taken when defining utility functions. It is feasible to represent the x-axis as the unrealised performance (i.e., $\bar{n}_g^1 - n_g^1$), instead of the output $n_g^1$. The following figure demonstrates this possibility. Hence, when unrealised output is low (i.e., $n_g^1$ is high), utility and achievement are high.

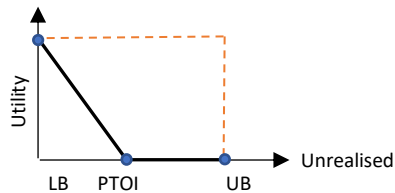

**Figure 4.** Alternative but equivalent utility function

### 3.3. Utility Functions - Financial

In CMP, financial considerations are equally important. Every patient treated in a public or private hospital has various costs associated with their care. Some of those costs may be met from the government, publicly funded universal healthcare organisations (i.e., like Medicare), and private health insurance. The rest may be incurred by the patient. Each patient is also charged and after costs are met, some hospitals may receive a net income/profit. When considering financial factors, it is perhaps warranted to define UF for each sub-group $p \in P_g$. It makes less sense by type, as significantly different costs/revenues occur at the level of subtype. If defined by subtype, the number of utility functions could be considerable.

Let us define $f_g$ as the net income received and $\gamma_g$ as the financial penalty incurred for each patient of type $g$. In 5(a) the simplest UF is posed. The utility increases linearly and is only zero when there is no output. The achievement is measured as the income generated and hence $u_g = n_g^1 f_g$ for $n_g^1 \leq UB_g$. For this UF, the reference point is $(\bar{n}_g^1, f_g \bar{n}_g^1)$, the breakpoints are $b_g = \{\bar{n}_g^1\}$ and the slopes are $\nabla_g = \{f_g, 0.0\}$.

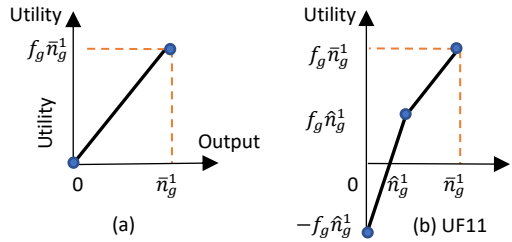

**Figure 5.** Financial utility functions (a) Without penalty, b) With penalty

If there is a demand, then it is possible to view unmet demand with regret and penalize the lost revenue. In Figure 5(b), there is a gain for $n_g^1 \geq \hat{n}_g^1/2$ and loss otherwise. The full income is only received if $n_g^1 \geq \hat{n}_g^1$. Partial gains are achieved in the range $[\hat{n}_g^1/2, d_g)$. Direct income is always $n_g^1 f_g$ and lost income of $(\hat{n}_g^1 - n_g^1)\gamma_g$ is subtracted when $n_g^1 < \hat{n}_g^1$. It is reasonable to set $\gamma_g = f_g$. The UF for this is as follows:

$$u_g = \begin{cases} n_g^1 f_g & n_g^1 \geq \hat{n}_g^1 \\ n_g^1 f_g - (\hat{n}_g^1 - n_g^1)\gamma_g & n_g^1 < \hat{n}_g^1 \end{cases} \quad (29)$$

For this UF, the reference point is $(\bar{n}_g^1, \bar{n}_g^1 f_g)$, the breakpoints are $b_g = \{\hat{n}_g^1\}$ and the slopes are $\nabla_g = \{2f_g, f_g\}$. The UF11 y-axis intercept is not -100. It changes relative to $\hat{n}_g^1$. If $\hat{n}_g^1$ is low then the y-intercept is not that small (i.e., only a little negative). If $\hat{n}_g^1$ is large, then regret is high, and the y-intercept has a very large negative value.

### 3.4. Utility Functions - Rewards and Regrets

In this section we consider unmet goals with regret and dissatisfaction and suggest further UF of practical relevance to CMP. Let us first consider the possibility that no utility, reward, or satisfaction is achieved if treatments for a particular group do not exceed a specified aspiration or demand $\hat{n}_g^1$. Otherwise, the utility increases linearly (or non-linearly) as $n_g^1$ increases. The UF for this is shown in Figure 6(a) and the UF for the linear variant is as follows:

$$u_g = \begin{cases} w_g n_g^1 & n_g^1 \geq \hat{n}_g^1 \\ 0 & n_g^1 < \hat{n}_g^1 \end{cases} \quad (30)$$

For this UF, the reference point is $(\bar{n}_g^1, w_g \bar{n}_g^1)$, the breakpoints are $b_g = \{\hat{n}_g^1, \hat{n}_g^1\}$ and the slopes are $\nabla_g = \{0.0, \hat{n}_g^1 w_g, w_g\}$. It is reasonable to set $w_g = f_g$ or $w_g = 1$. For the first option, the reward is monetary and specific to a particular patient group. The second option is independent of group and non-monetary. Another option is to set $w_g = 100/\bar{n}_g^1$. The utility can then be viewed as a level of satisfaction between 0 and 100%.

Let us then consider the loss of reward, income, and satisfaction from not meeting demand. We could penalize the extent of the unmet demand and include that value as a measure of dissatisfaction. This is shown in Figure 6(b). Evidently, outputs below demand are assumed to provide no satisfaction here. The UF for this is as follows:

$$u_g = \begin{cases} w_g n_g^1 & n_g^1 \geq \hat{n}_g^1 \\ -\gamma_g(\hat{n}_g^1 - n_g^1) & n_g^1 < \hat{n}_g^1 \end{cases} \qquad (31)$$

For this UF, the reference point is $(\bar{n}_g^1, w_g \bar{n}_g^1)$, the breakpoints are $b_g = \{\hat{n}_g^1, \hat{n}_g^1\}$ and the slopes are $\nabla_g = \{\gamma_g, w_g \hat{n}_g^1, w_g\}$. The parameter $w_g$ can be defined in three ways as previously mentioned, and as such it makes sense to define $\gamma_g = w_g$ as well.

Third, it is worth considering only regret and not the measurement of reward. Figure 6(c) shows a UF modelling only regret. In that function, any output above demand is assumed to have no utility. The UF for this is as follows:

$$u_g = \begin{cases} 0 & n_g^1 \geq \hat{n}_g^1 \\ -(\hat{n}_g^1 - n_g^1)\gamma_g & n_g^1 < \hat{n}_g^1 \end{cases} \qquad (32)$$

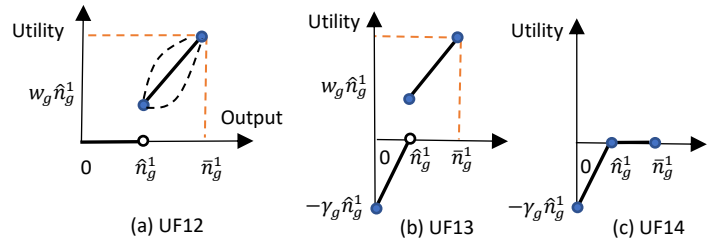

(a) UF12  (b) UF13  (c) UF14

**Figure 6.** UF measuring satisfaction and regret

**Final Remarks**. The UF shown in Figure 6 are basic as they have only two segments. More segments can be added on both sides of the discontinuity and on non-static segments. The UF shown in Figure 6(a) is well suited to current funding arrangements in Australia, whereby public hospitals are given extra funding for exceeding planned "care targets".

Defining valid UF is a key step to performing CMP. The process of eliciting UF, identifies the objectives of managers, and places limitations on the possible solutions that can be entertained. This reduces the complexity of the optimisation and calculations phase. As reported in the literature, it is likely that an iterative process is needed to provide/revise these, as extracted UF may be contentious. Between each stage of that process, an analysis would be performed followed by a negotiation. As UF are needed for each patient type or group, it is foreseeable that many stakeholders would need to be approached to gather a full set of UF. When eliciting a UF, each stakeholder may be asked questions like the following:

*Question 1*. Which metric is being used to measure performance? For instance, is the function modelling satisfaction/dissatisfaction, profit/loss, cost, achievement/non-achievement?
*Question 2*. Is there a minimum or maximum output? If there is, what is the value? Are there thresholds of acceptable performance, below which a DM will not be prepared to go, no matter what the gains in other criteria?
*Question 3*. Is there an aspiration, target, or demand for each group?

*Question 4*. For what range of values is the utility function deemed "static", "increasing" or "decreasing"? Is the increase concave up (⌣), concave down (⌢), or linear? Is the decrease concave up (⌣), concave down (⌢), or linear?

*Question 5*. Is over and under-achievement undesirable? Is there a penalty for not meeting the minimum output or for exceeding the maximum output? What is the meaning of the penalty and what is the penalty value?

*Question 6*. Regarding the definition of discontinuous and tiered functions, are the boundaries of linear segments open or closed?

**4. Case Study**

**4.1. Details**

In this case study we have considered a large tertiary level public hospital in the local area. For the purposes of our CMP activities, the main infrastructure of the hospital has been included, for instance a 26-bed intensive care unit, 19 operating theatres, and 24 surgical/medical wards totalling 522 beds. Excluded from further consideration are surgical care areas for preoperative and post anaesthesia care and some miscellaneous wards. The time availability of wards is 168 hours per week, and 40 hours for operating theatres.

Historical patient treatment information has been collected and this constitutes the main inputs to the CMP. From the historical data, patient treatment times and resource demands have been extracted. Within each specialty, there are many subtypes. These are characterised by Diagnosis Related Group (DRG), a classification system that groups hospital cases by the resources required in their treatment. The number of DRGs however is prohibitive, so for pragmatic reasons we have characterised patients as either surgical or medical inpatients in this article. An unrestricted case study with subtypes defined by DRG can be found in Burdett et al. (2023a).

For the 19 specialties we have chosen to consider, and for each inpatient subtype, there is an average time requirement for, i) surgery in an operating theatre, ii) recovery or other treatment in a ward, and iii) intensive care. These times are weighted averages scaled by the prevalence of each DRG. Table 2 describes the considered specialties and the number of subtypes. The TRANS patient type only includes surgical inpatients, and the PSY type only includes medical inpatients. Historically, the records showed that medical inpatients also required intensive care and surgeries. Table 3 describes the wards and their focus, i.e., the types of patients that should be cared for there.

**Table 2.** Patient types, subtypes, and resourcing details

| # | SPECIALTY (GROUP) | #SUB | (SUR, MED) %MIX | SURGICAL (OT, ICU, WARD) AVG TIME (HRS) | WARDS | MEDICAL (OT, ICU, WARD) AVG TIME (HRS) | WARDS |
|---|---|---|---|---|---|---|---|
| 1 | Cardiac (CARD) | 2 | (58.8,41.2) | (3.16,19.85,171.35) | 3C | (0.06,1.82,84.45) | 3D, 3E, 5A |
| 2 | Endocrinology (ENDO) | 2 | (50.63,49.37) | (2.13,2.72,137.85) | 4D | (0.51,0.27,185.24) | 4D, 5C |
| 3 | Ear Nose Throat (ENT) | 2 | (54.08,45.92) | (2.12,1.02,44.02) | 1D | (0.5,0.91,49.43) | 1D |
| 4 | Facio-Maxillary (FMAX) | 2 | (70.67,29.33) | (4.52,6,131.33) | 1D | (0.61,0.08,13.55) | 1D |
| 5 | Gastroenterology (GAST) | 2 | (54.97,45.03) | (2.64,3.61,150.71) | 4D, 4E | (0.144,0.49,101.43) | 4D, 4E, 5C |
| 6 | Gynaecology (GYN) | 2 | (67.45,32.55) | (2.2,10.4,111.36) | 4C | (0.59,0,52.86) | 4C, 5C |
| 7 | Hepatology (HEP) | 2 | (45.97,54.03) | (1.475,4.13,160.71) | 4C, 4E | (0.075,1.84,119.87) | 4C, 4E |
| 8 | Immunology (IMMU) | 2 | (5.66,94.34) | (1.93,4.3,306.79) | 2D | (0.19,44.68,149.15) | 2D, 5B |
| 9 | Nephrology (NEPH) | 2 | (28.3,71.7) | (2.19,0.65,102.41) | 4BR | (0.47,0.143,50.65) | 4BR, RENDP, 5C |
| 10 | Neurology (NEUR) | 2 | (26.95,73.0) | (2.46,3.67,243.44) | 2C | (0.099,5.35,200.68) | 2C, 5B |
| 11 | Oncology (ONC) | 2 | (57.28,42.72) | (2.86,2.09,217.5) | 2E | (0.36,0.89,172.27) | 2E |
| 12 | Ophthalmology (OPHT) | 2 | (68.83,31.17) | (1.52,0.068,45.35) | 4D | (0.046,0,100.36) | 5A |
| 13 | Orthopaedics (ORTH) | 2 | (64,36) | (3.09,1.93,218.98) | 2A, 2B | (0.52,1.86,266.12) | 2A, 2B |
| 14 | Plastics (PLAS) | 2 | (65.69,34.31) | (2.43,1.71,157.44) | 1D | (0.18,0.1,137.73) | 1D |
| 15 | Psychiatry (PSY) | 1 | (100) | na | na | (0.08,0.06,258.82) | GREV |
| 16 | Respiratory (RESP) | 2 | (5.62,94.38) | (2.86,3.7,161.26) | 2D | (0.22,4.76,136.37) | 2D, 5A |
| 17 | Transplants (TRANS) | 1 | (100) | (3.33,445.71,593.24) | 4BT | na | na |
| 18 | Urology (UROL) | 2 | (43.73,56.27) | (1.83,1.66,71.63) | 4A | (0.38, 0.1, 41.11) | 4A |
| 19 | Vascular (VASC) | 2 | (31.85,68.15) | (2.98,4.75,339.59) | 1C | (0.07,5.9,122.74) | 1C |

**Table 3.** Ward information

| # | WARD | #BEDS | FOCUS | # | WARD | #BEDS | FOCUS |
|---|------|-------|-------|---|------|-------|-------|
| 1 | 1C | 24 | VASC | 12 | 4BR | 14 | NEPH |
| 2 | 1D | 26 | ENT, FMAX, PLAS | 15 | 4BT | 16 | TRANS |
| 3 | 2A | 28 | ORTH | 16 | 4C | 28 | HEP, GYN |
| 4 | 2B | 26 | ORTH | 17 | 4D | 28 | GAST, ENDO, OPHT |
| 5 | 2C | 36 | NEUR, ONC | 18 | 4E | 26 | GAST, HEP, GYN |
| 6 | 2D | 24 | IMMU, RESP | 19 | 5A | 28 | OPHT(MED), CARD(MED), RESP(MED) |
| 7 | 2E | 29 | ONC | 20 | 5B | 24 | IMMU(MED), NEUR (MED) |
| 8 | 3C | 28 | CARD (SUR) | 21 | 5C | 24 | ENDO(MED), GAST(MED), NEPH (MED) |
| 9 | 3D | 20 | CARD (MED) | 22 | 5D | 24 | INFD |
| 10 | 3E | 14 | CARD (MED) | 23 | RENDP | 6 | NEPH (MED) |
| 11 | 4A | 19 | UROL | 24 | GREV | 30 | PYS |

The maximum number of patients treatable in each group (i.e., $\bar{n}_g^1$) to the detriment of others have been computed by applying the core CMP model of section 3.1. The results are shown in Table 4.

**Table 4.** Patient treatment limits (a.k.a., bounds)

| GROUP | CARD | ENDO | ENT | FMAX | GAST | GYN | HEPA | IMMU | NEPH | NEUR |
|-------|------|------|-----|------|------|-----|------|------|------|------|
| UB ($\bar{n}_g^1$) | 2427.78 | 2817.25 | 4884.2 | 1820.53 | 5301.99 | 5109.98 | 3261.53 | 2652.76 | 4219.99 | 2470.08 |
| **GROUP** | **ONC** | **OPHT** | **ORTH** | **PLAS** | **PSY** | **RESP** | **TRANS** | **UROL** | **VASC** | **TOTAL** |
| UB ($\bar{n}_g^1$) | 1278.37 | 6083.21 | 1999.34 | 1507.43 | 1012.60 | 3297.35 | 235.61 | 3048.02 | 649.9 | 54077.91 |

## 4.2. Sensitivity Analysis

It is necessary to map the behaviour of different UF within the context of CMP. To the best of our knowledge utility functions are not currently used in hospitals and as such we cannot test the preferences and aspirations of real hospital managers. Hence, we analyse each UF type individually. This aligns with a hospital wide alignment of values and beliefs regarding outputs. However, different parameters of each UF type are considered.

The purpose of the numerical testing is to find caseloads with highest utility, and this may translate on some occasions to caseloads with a higher or lower number of patients treated. Our numerical results are summarised in Appendix D and includes most of the UF discussed in section 3. The piecewise linear functions of each UF can also be found in Appendix B and C. To characterise the piecewise linear functions of all non-linear UF, thirty break points were arbitrarily selected. For the UFM there are two parameters in the objective, namely $\epsilon_1$ and $\epsilon_2$. We considered the two extremes, namely $\epsilon_1 = 1, \epsilon_2 = 0$ and $\epsilon_1 = 0, \epsilon_2 = 1$. The former maximizes the minimum utility (MMU), and the later maximizes the sum of the utilities (MSU).

**GAM and GPM.** The GAM and GPM were applied for comparative purposes. Their details are summarised in Table 5, where $u_g = 100 n_g^1 / \bar{n}_g^1$. The GAM method has identified the least equitable caseload of the two methods but has higher outputs for a select number of groups. The GAM solution has nine groups with zero output. The GPM (MSU) has two groups at zero output and a third that is close to zero. The GPM (MMU) however achieves consistent outputs of 36% for each group except CARD and TRANS, where output is at 100%. The total number treated, however, is lowest.

**Table 5.** Results of traditional GSM and GPM approaches

| Type | $\mathbb{N}$ | $\sum_g u_g$ | $\min_g u_g$ | $\text{avg}_g(u_g)$ | $\max_g u_g$ |
|------|---|---|---|---|---|
| GAM | 22389.66 | 513.55 | 0 | 27.03 | 100 |
| GPM (MMU) | 21232.76 | 813.36 | 36.03 | 42.81 | 100 |
| GPM (MSU) | 31663.97 | 1325.00 | 0 | 69.74 | 100 |

**UF1** . The results for UF1 are summarised in Figure 7 and 8. The numerical testing clearly shows that the convexity (dictated by parameter $\alpha$) has a significant effect on the parameters of interest, namely $\mathbb{N}, \sum_g u_g$ and $\min_g u_g$. Highest aggregate utility and highest minimum utility occur when convexity of UF1 is downwards. For the MSU objective, there is a difference of 2781 patients per year, between the non-linear variants. The convex up version produces the lowest number of patients treated whereas the convex down version the highest. For the MMU objective, however, the largest number of patients is when there is no convexity (i.e., the UF is linear).

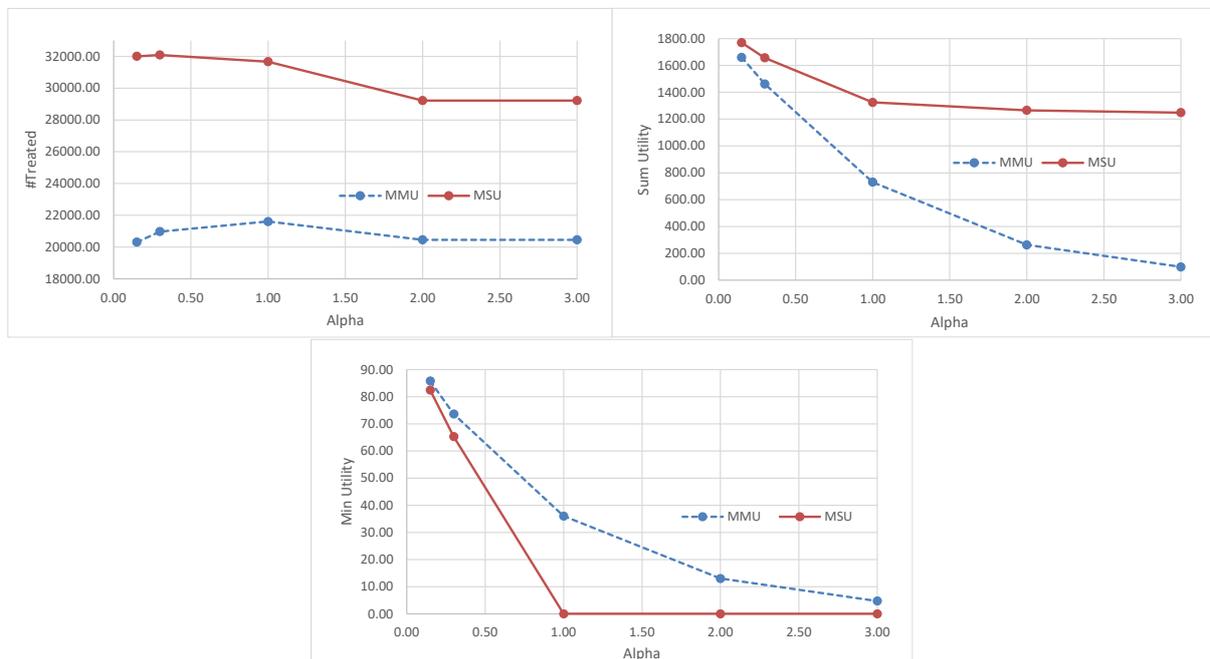

**Figure 7**. Alpha versus total treatments, sum of utility, and minimum utility for UF1

The caseloads obtained are significantly different between the MMU and MSU cases, across different values of $\alpha$. The difference between the maximum and minimum of each group are summarised in Figure 8. That chart summarises the prevailing case mix, which is computed as $100 n_g^1 / \mathbb{N}$. That chart shows much less variation in the MMU objective. Specialty GAST, NEPH and PSY were altered the most. For the MSU objective, more groups were altered.

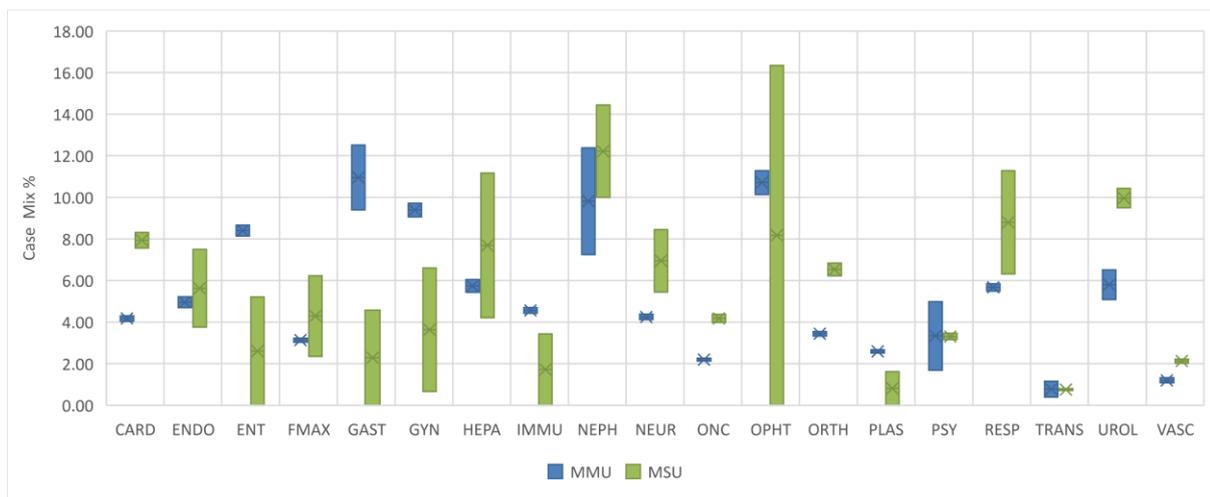

**Figure 8.** Differences observed in the case mix (UF1)

**UF2** ⬜. The results for UF2 are summarised in Figure 9 and 10. The maximum number of treatments are obtained when there is no indifference. When the point of indifference is increased, $\mathbb{N}$ decreases slowly for the MSU objective and quickly for the MMU objective. When the level of indifference becomes too high, it is not possible for each group to have non-zero utility, and a "zero" solution is produced. For the MMU case, ℕ drops to zero when the indifference level is 40%. This means that no single group has an output above the level of indifference. For the MSU case, trade-offs are made straightaway to achieve higher outputs, and three groups have a zero utility. Also, there is no solution for a 100% level of indifference. The difference between the maximum and minimum of each group are summarised in Figure 10. There are significant changes required in some specialties.

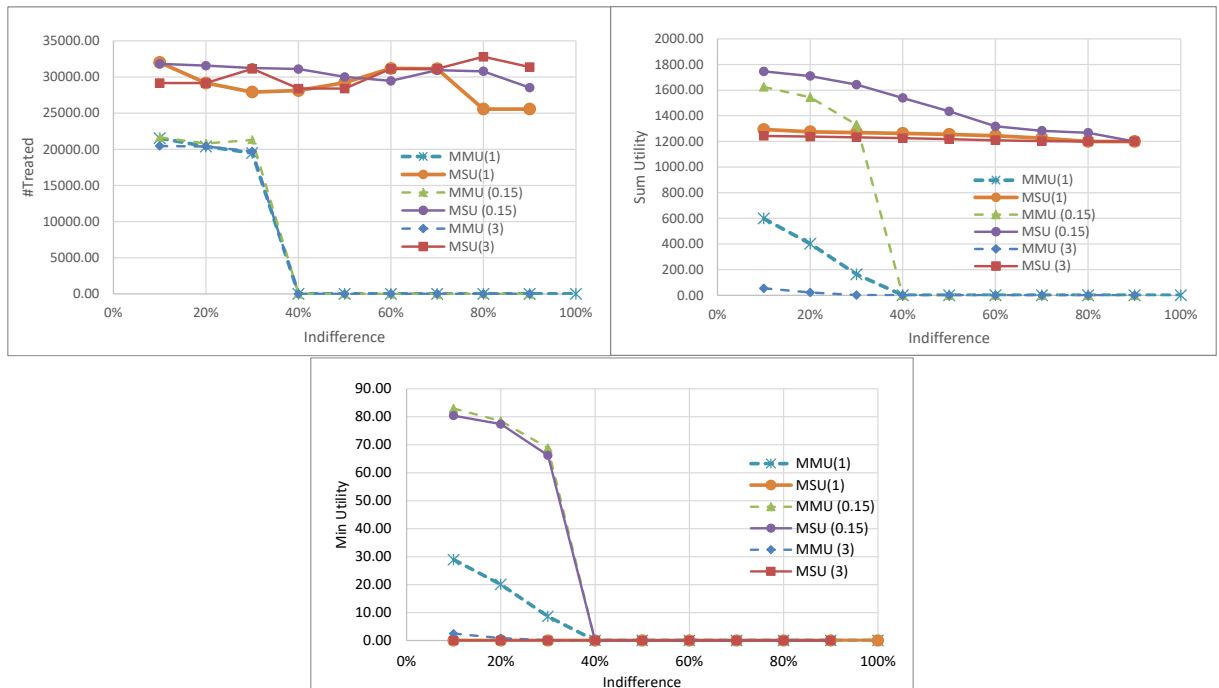

**Figure 9**. Indifference versus total treatments, sum of utility, and minimum utility for UF2

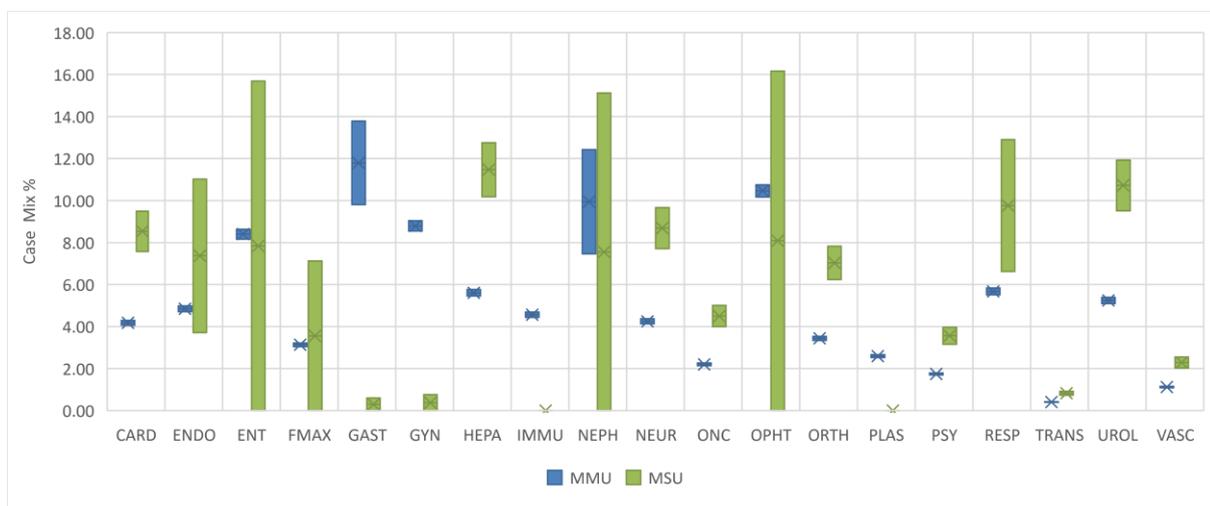

**Figure 10.** Differences observed in the case mix (UF2) when $\alpha = 1$

**UF3**. The results for UF3 are summarised in Figure 11 and 12. The numerical testing shows that if group aspiration is assigned low, then ℕ will be low, and total utility $\sum_g u_g$ will be high. As aspiration is increased, then ℕ increases, and total utility decreases. Total utility decreases, because as more system capacity is consumed, the aspiration level of each group may not be achievable, i.e., trade-offs need to be made. The MMU case is most affected by the aspiration level. In our case study, when the aspiration level is greater than 36% of capacity, trade-offs need to be made. Prior to that, all groups could meet the specified aspiration level. There is little deviation in the MMU case mix. There are more significant changes in the MSU case mix, however. Increased percentages are achieved at the expense of reductions in ENT, GAST, GYN, NEPH and OPHT. The effect of non-linearity is quite pronounced in each measure.

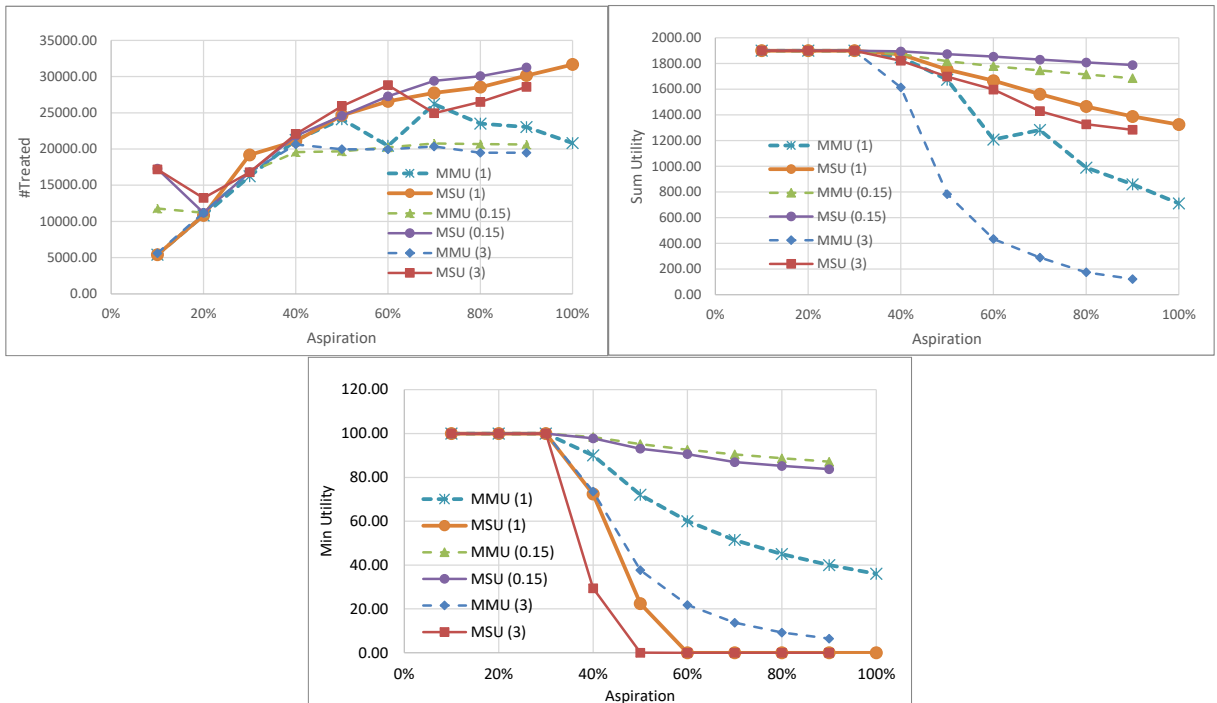

**Figure 11**. Aspiration level versus total treatments, sum of utility, and minimum utility for UF3

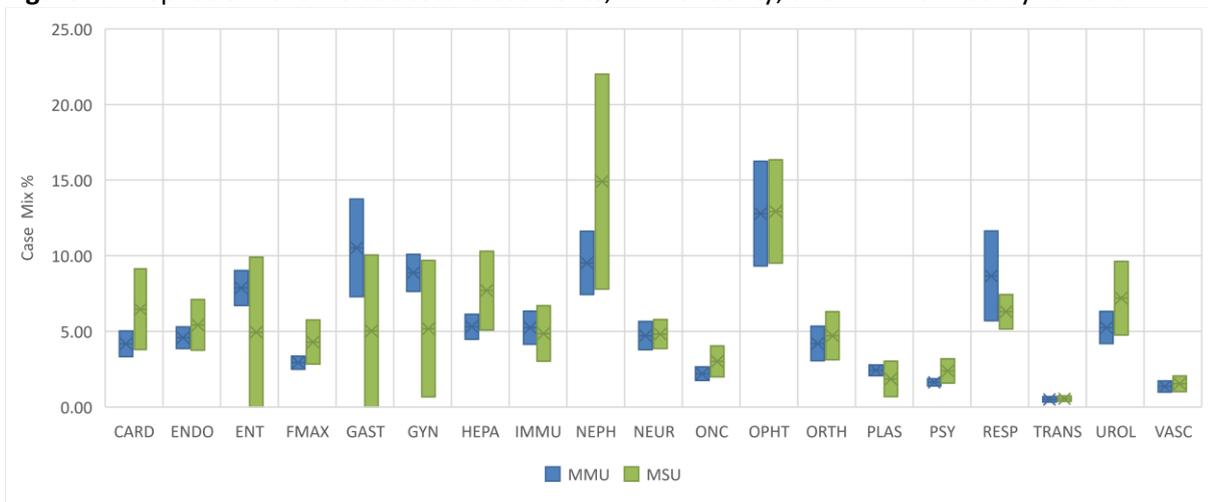

**Figure 12.** Differences observed in the case mix (UF3) when $\alpha = 1$

**UF4** ⌐⌐. The results for UF4 are summarised in Figure 13 and 14. The numerical testing shows that $\mathbb{N}$ does not change much as the indifference and aspiration level are altered. When the indifference level is high enough, however, the MMU case does produce a zero solution. As the indifference level and aspiration approach 50% symmetrically from below and above respectively, total utility tends to increase for the MSU case, but the opposite for the MMU case. The MSU case has at least one zeroed group, and hence a minimum utility of zero always. For the MMU case, the minimum utility decreases slowly.

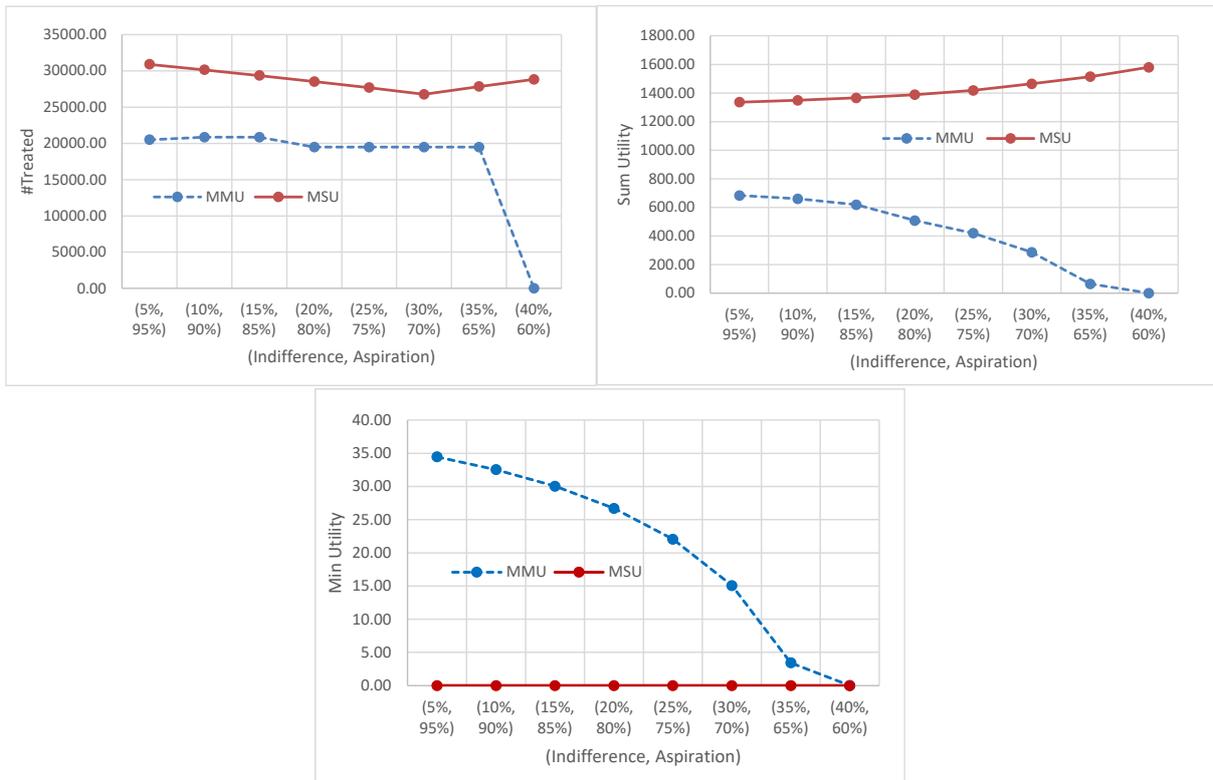

**Figure 13**. Total treatments, sum of utility, and minimum utility for UF4

As the parameters are altered the same caseload and case mix are obtained for the MMU case; the only variation occurs in a few specialties like GYN. For the MSU case, the percentage decreases for some specialities but increases for others. This is not shown in Figure 14 but is clearly visible in a standard bar-chart of the caseload.

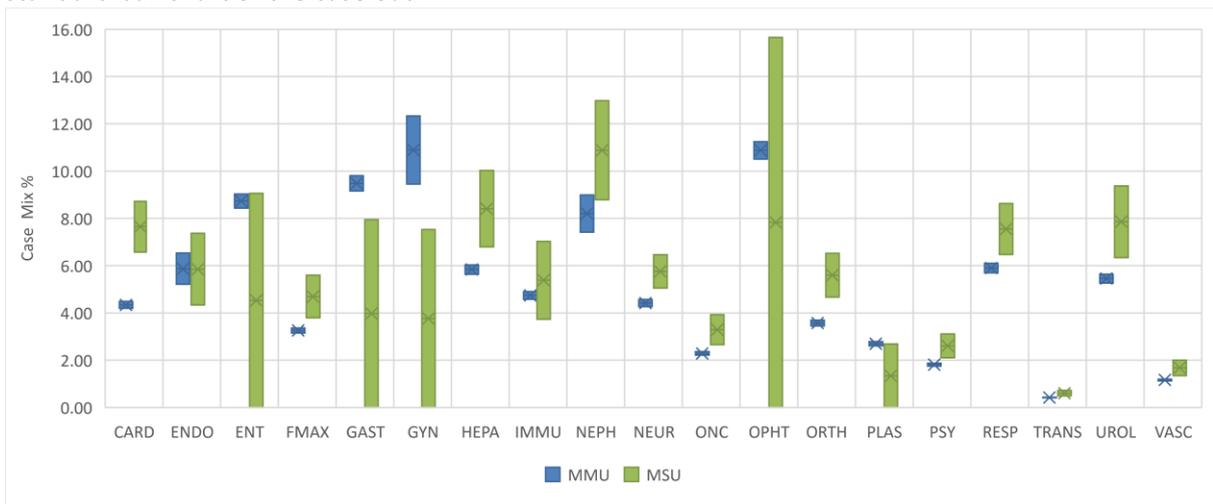

**Figure 14.** Differences observed in the case mix (UF4)

**UF5** . The results for UF5 are summarised in Figure 15 and 16. When the intercept is high, the utility at zero is very small, in fact smaller than -100. However, the utility is always 100 at the upper bound. We can see that $\mathbb{N}$ does not change at all as the intercept is increased. However, a decrease in total utility occurs. This decrease is somewhat linear over the % range (0,50], but significantly more non-linear when over 60%. In this situation, the reduced total utility is not an indicator that trade-offs are being made as the intercept is altered. Scrutiny of the results shows that the same caseload is obtained (regardless) for the MSU case. For the MMU case, the caseloads are very similar too, but there are some significant changes in ENDO, GAST, GYN and NEPH. The reduced utility eventuates not because the $n_g^1$ values change but rather because the slope of the function steepens as the intercept increases.

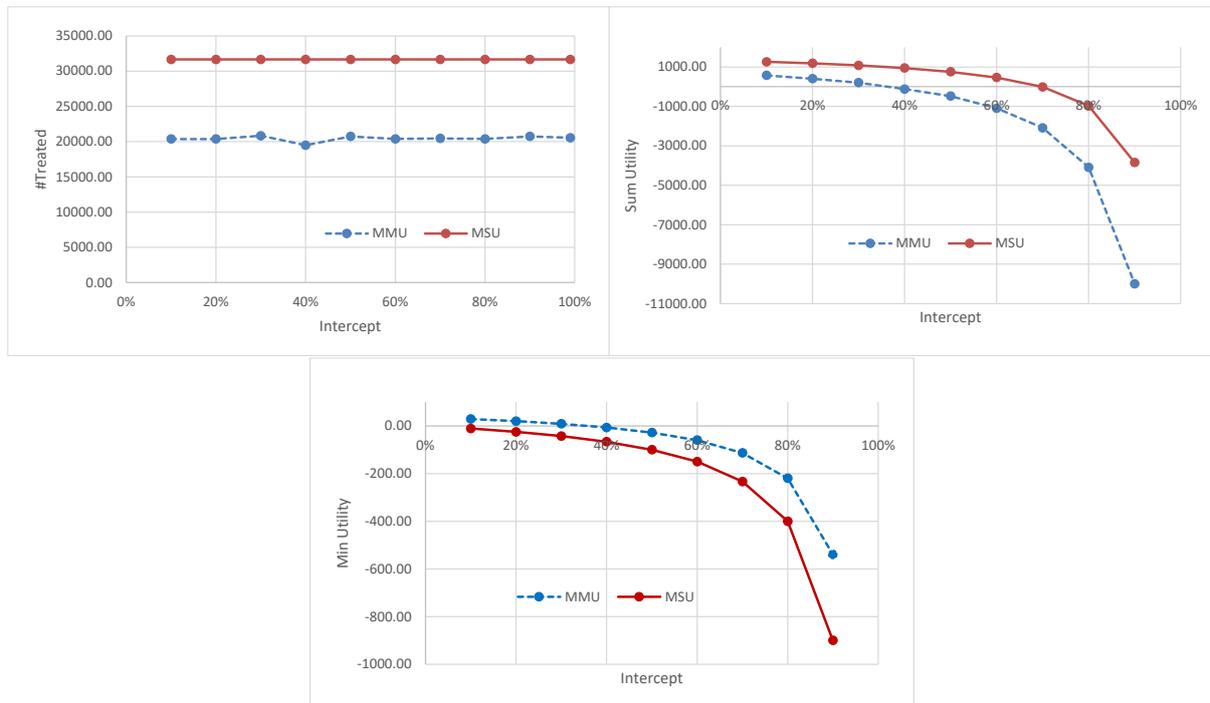

**Figure 15**. Aspiration level versus total treatments, sum of utility, and minimum utility for UF5

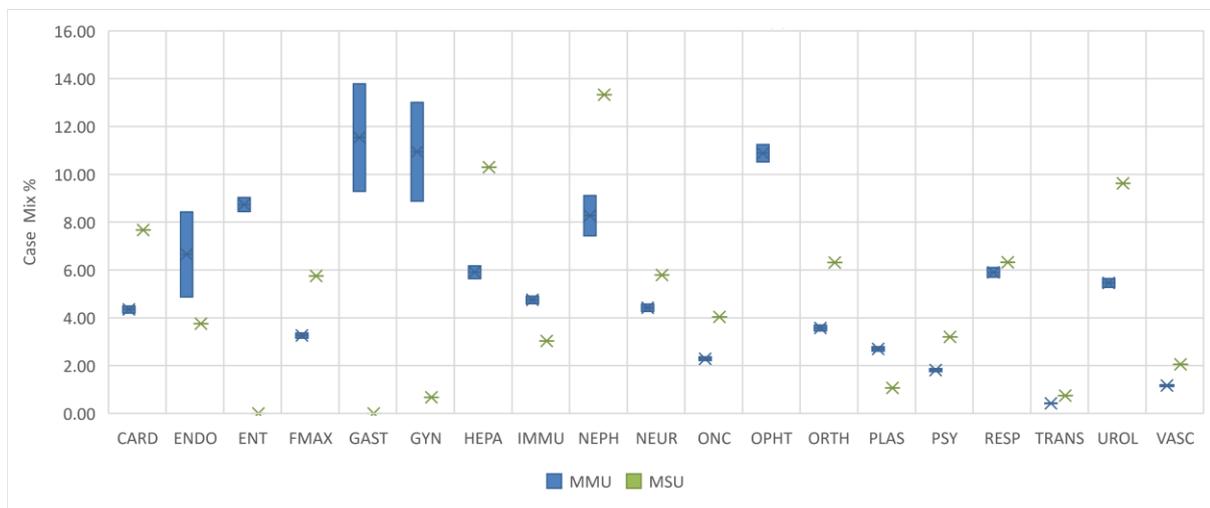

**Figure 16.** Differences observed in the case mix (UF5)

**UF6** . The results for UF6 are the same as those obtained for UF3. There are, however, some differences in the case mix as shown in Figure 17, relative to Figure 12. As UF6 has two values of $n_g^1$ for each utility value, the CMP economizes and chooses values on the left-hand side of the apex first. Without a term involving $\mathbb{N}$ in the objective, outputs $n_g^1 > n_g^A$ will not occur, because the utility decreases on the right-hand side of the apex.

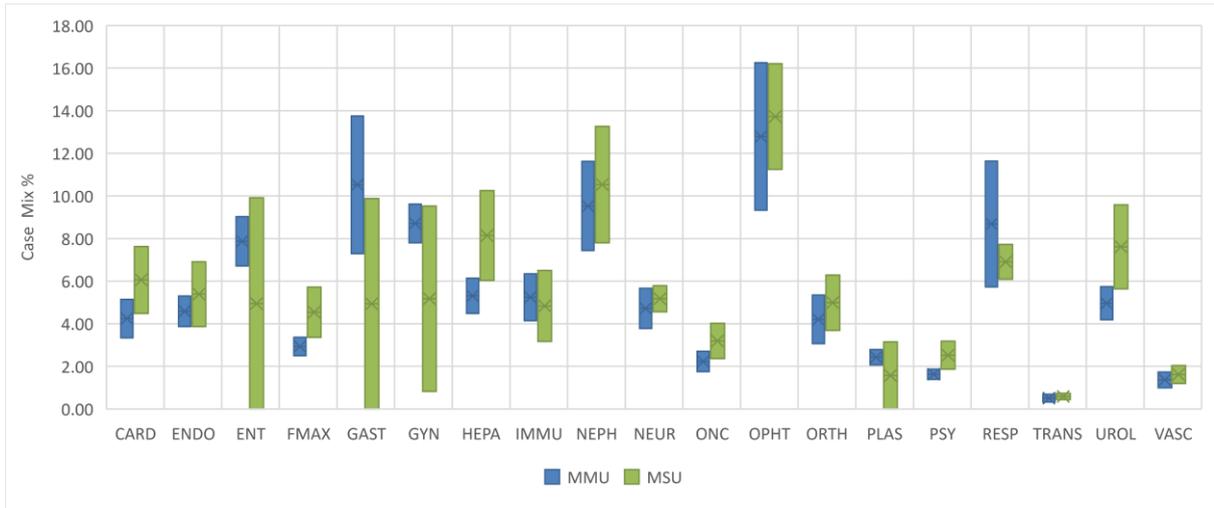

**Figure 17.** Differences observed in the case mix (UF6)

**UF7** ⌐_⌐. The results for UF7 are summarised in Figure 18 and 19. Although UF7 is like UF4, the results are unalike. As the reference point is increased and the s-shape is shifted, there is little change in total number of treatments, but significant decreases in total utility and minimum utility. The case mix for the MMU objective is quite invariant. For the MSU objective, the case mix varies considerably as the reference point is increased. Quite a few specialties are zeroed.

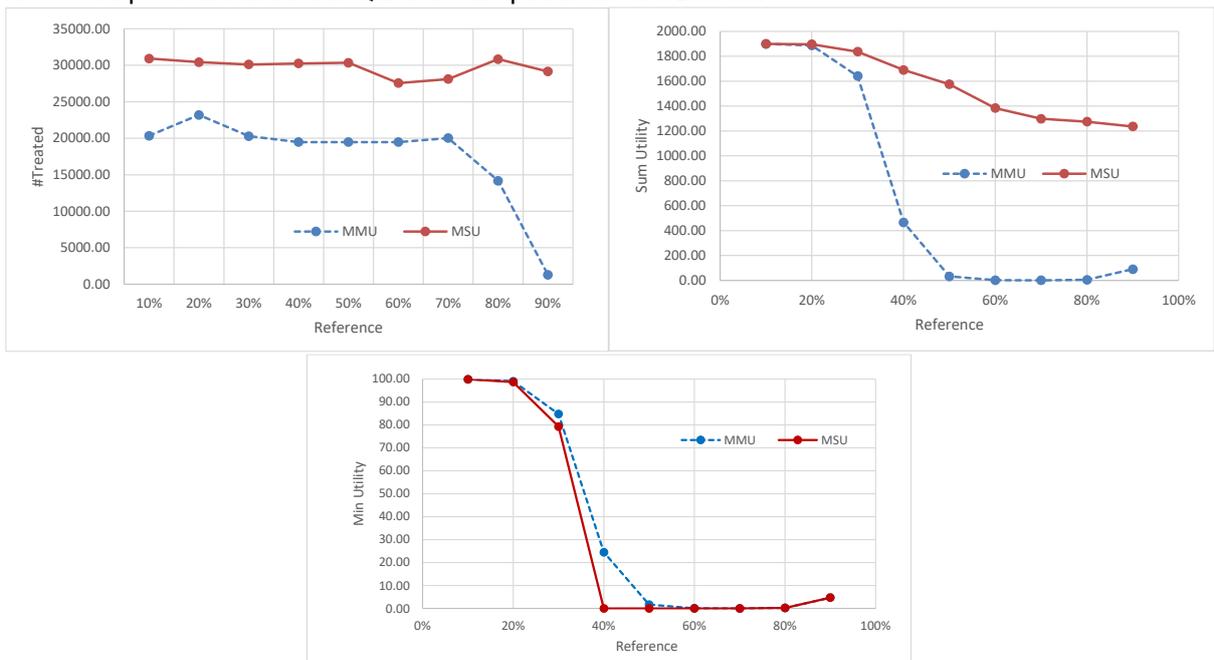

**Figure 18**. Reference point versus total treatments, sum of utility, and minimum utility for UF7

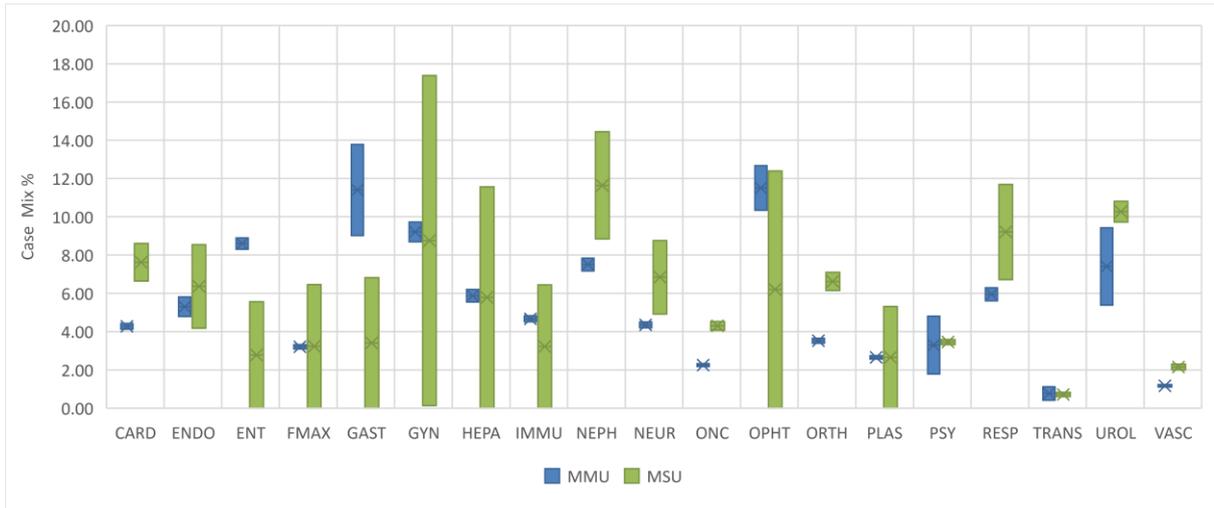

**Figure 19.** Differences observed in the case mix (UF7)

**UF8** ▭. The results for UF8 are summarised in Figure 20 and 21. This UF has similar behaviour to UF3 initially. However, notable differences arise as the level of indifference is increased. Once it is too high, a non-zero solution is not achievable for the MMU objective. Solutions are however achievable for the MSU objective as some groups can be zeroed to allow other groups to prosper. The total number of patients treated has not increased uniformly and perhaps demonstrates the possibility of finding alternatively optimal caseloads. The total utility however gradually decreases. The case mix is quite variable for the MSU objective, perhaps the most of any of the UF. There are big differences between the min and max percentage. This occurs because UF8 is tiered, and there are only two utility values, namely 0 and 100. As such, there are many alternatively optimal caseloads to choose from.

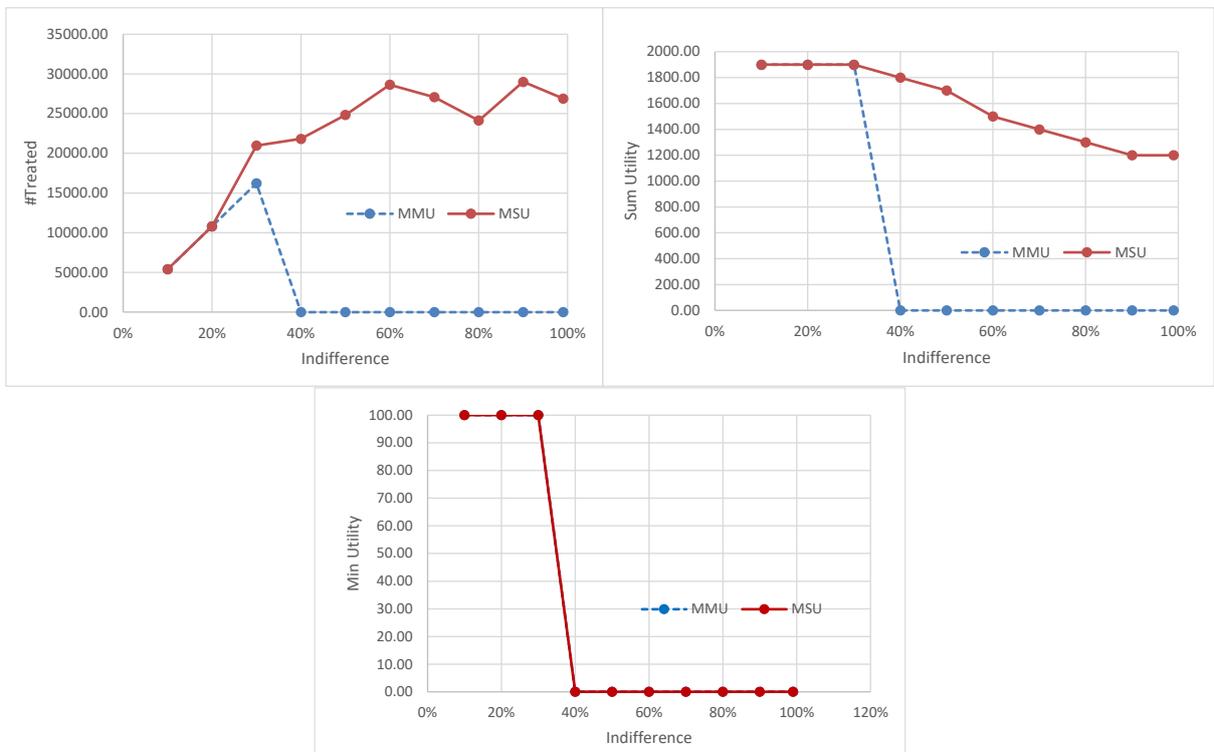

**Figure 20**. Indifference versus total treatments, sum of utility, and minimum utility for UF8

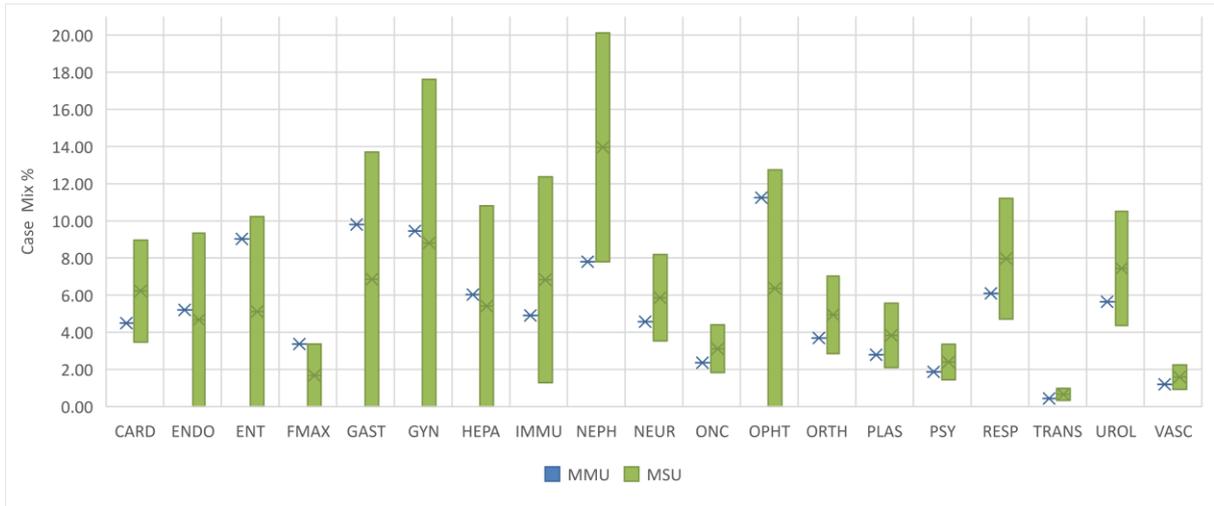

**Figure 21.** Differences observed in the case mix (UF8)

**UF11** . The results for UF11 are summarised in Figure 22 and 23. The numerical testing shows that the results are quite comparable to those of UF5, and this makes sense as they have a common segment with negative utility. It is worth noting that when goals are high, regret is highest and negative utilities are incurred for some groups. At most, six specialties had negative utility, and this occurred when indifference was at 90% of specialty capacity. When goals are low, regrets are small.

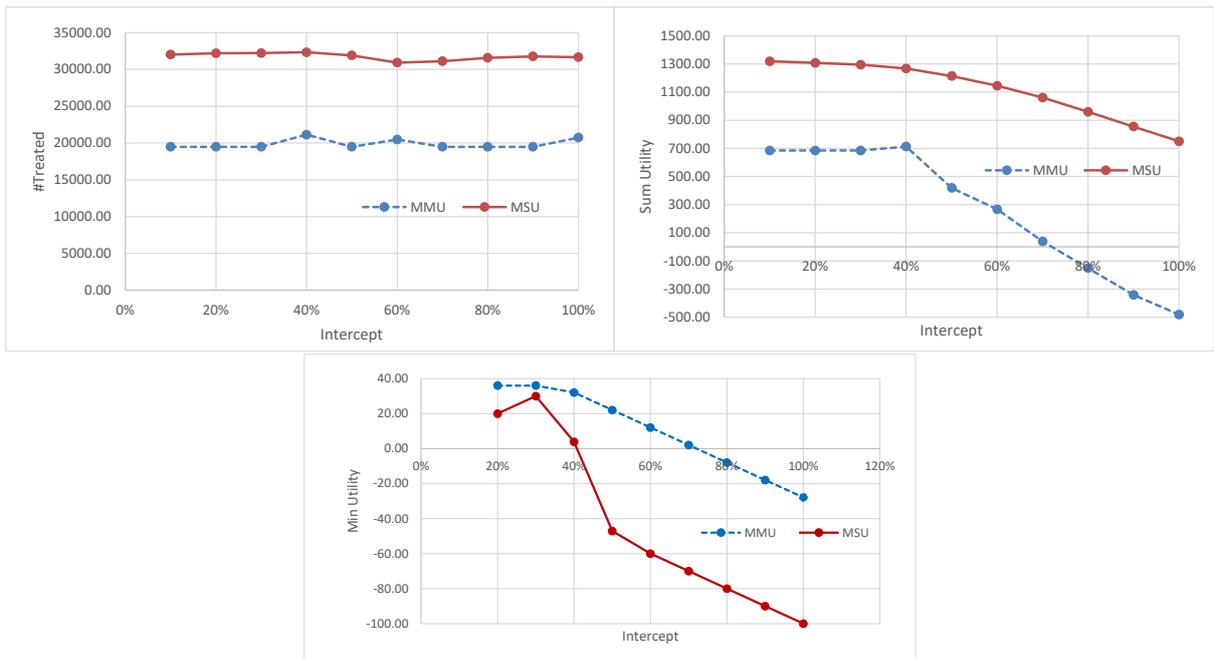

**Figure 22.** Differences observed in the case mix (UF11)

The case mix varies only slightly for the MMU objective, and quite a lot for the MSU objective. Interestingly some specialities had almost no variation at all in the different caseloads, like CARD, ONC, ORTH, PSY, TRANS, VASC.

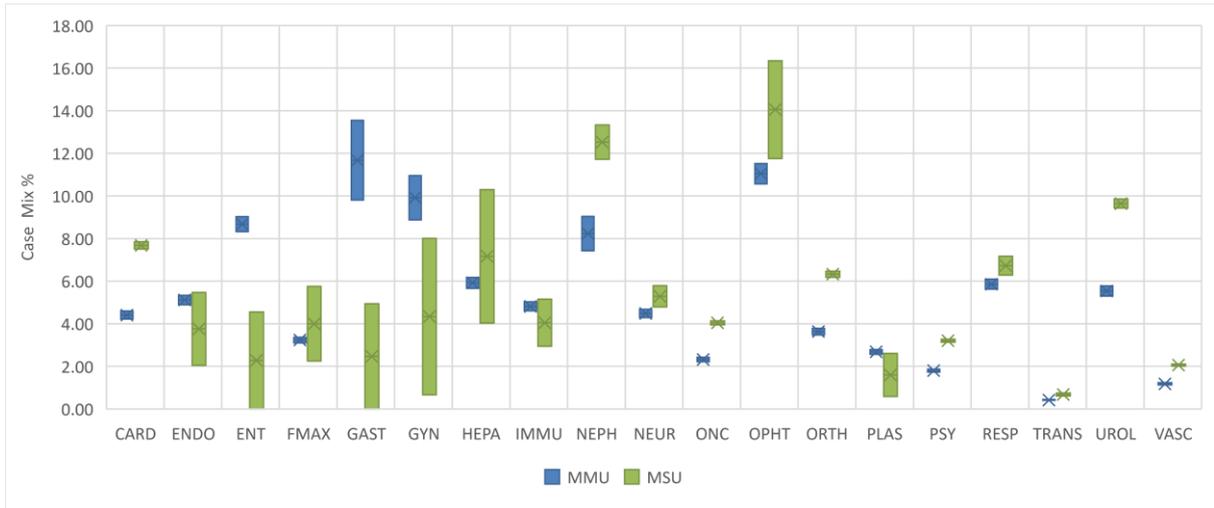

**Figure 23.** Differences observed in the case mix (UF11)

**UF12** . The results for UF12 are summarised in Figure 24 and 25. Although UF8 and UF12 are the same up till the point of indifference, the results are not completely alike. When the level of indifference is small, then it is unnecessary for any group to have a zero utility; the utility of each group will lie on the slope somewhere. If the level of indifference is larger, some groups will be zeroed and the rest will have a utility on the slope, further up, away from the point of discontinuity. As the indifference increases, more groups will occur at the point of discontinuity, and the majority will need to be zeroed.

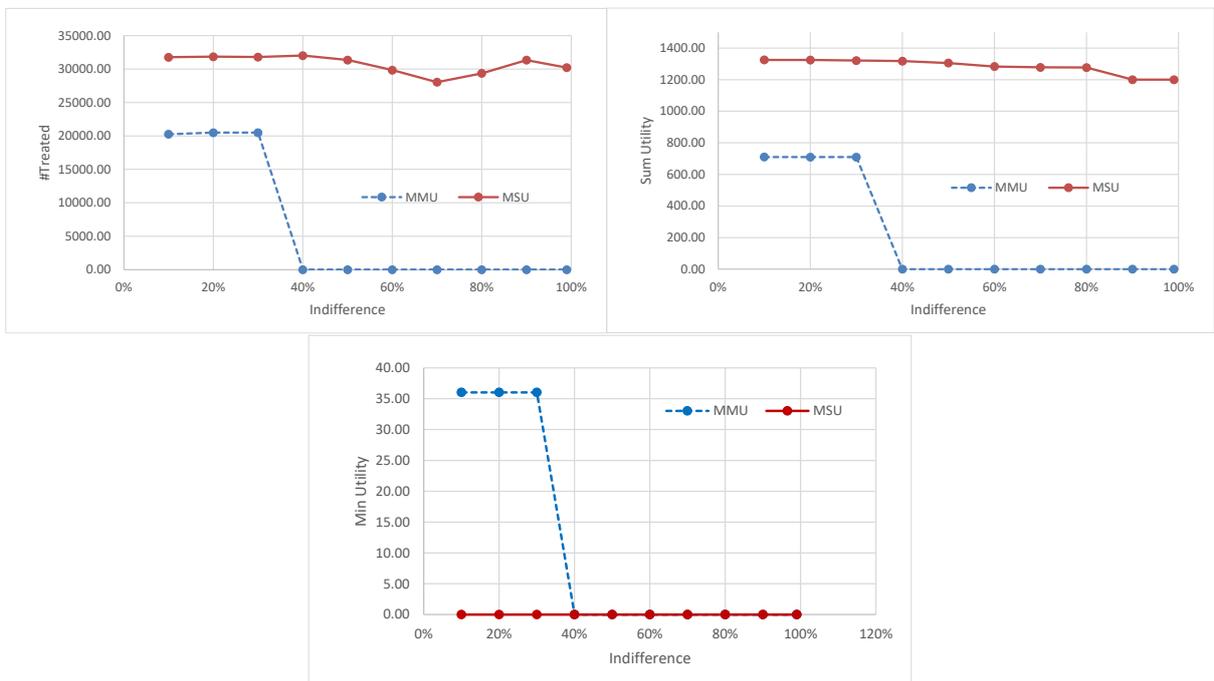

**Figure 24**. Indifference versus total treatments, sum of utility, and minimum utility for UF12

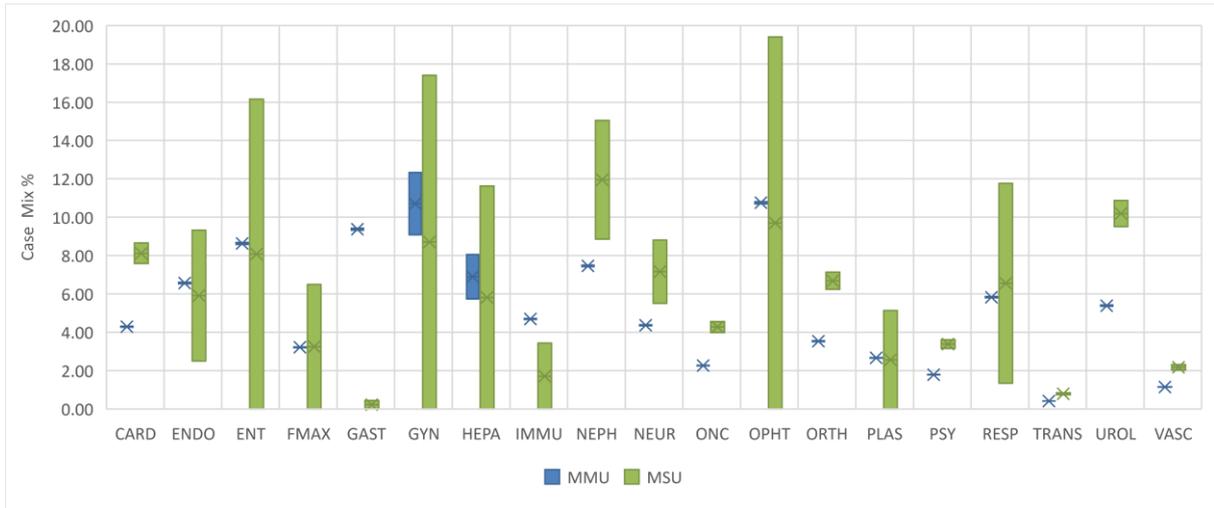

**Figure 25.** Differences observed in the case mix (UF12)

**UF13** . The results for UF13 are summarised in Figure 26 and 27. These are like those of UF12. There are however two main differences. For the MSU objective, the total utility decreases more significantly. Also, the minimum utility is negative, as opposed to zero. The MMU case mix results are invariant, except for a percent difference in ENDO and GYN. For the MSU objective, six specialties have an unvarying case mix.

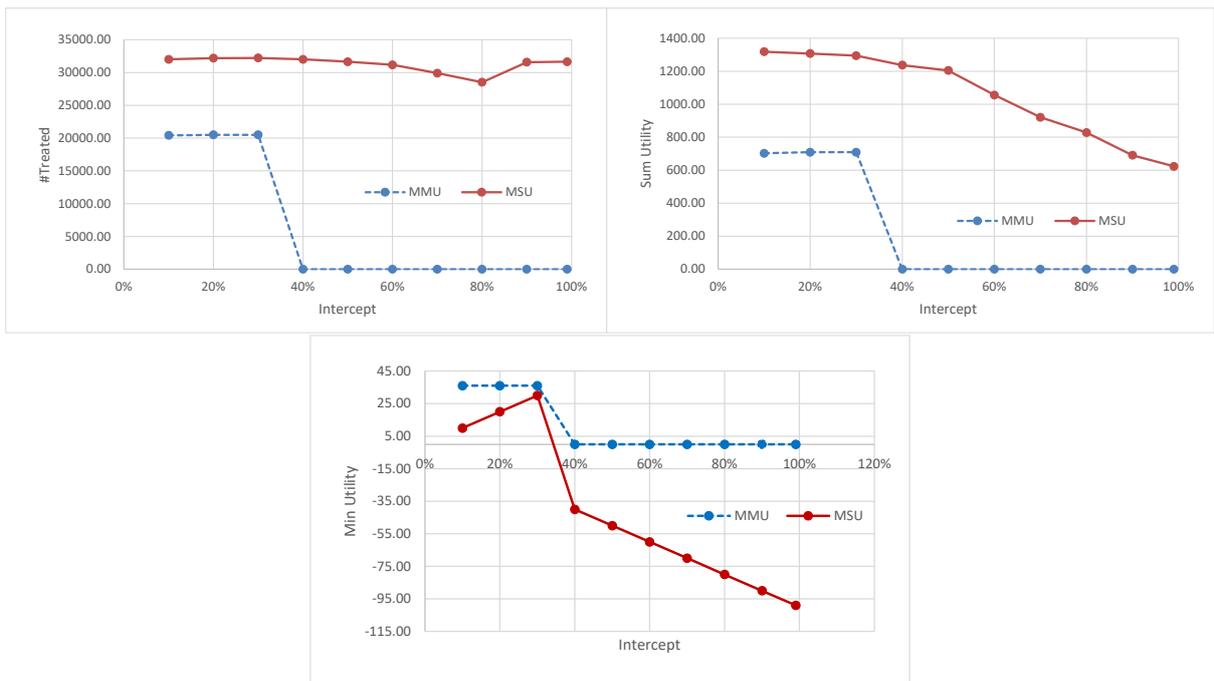

**Figure 26**. Intercept versus total treatments, sum of utility, and minimum utility for UF13

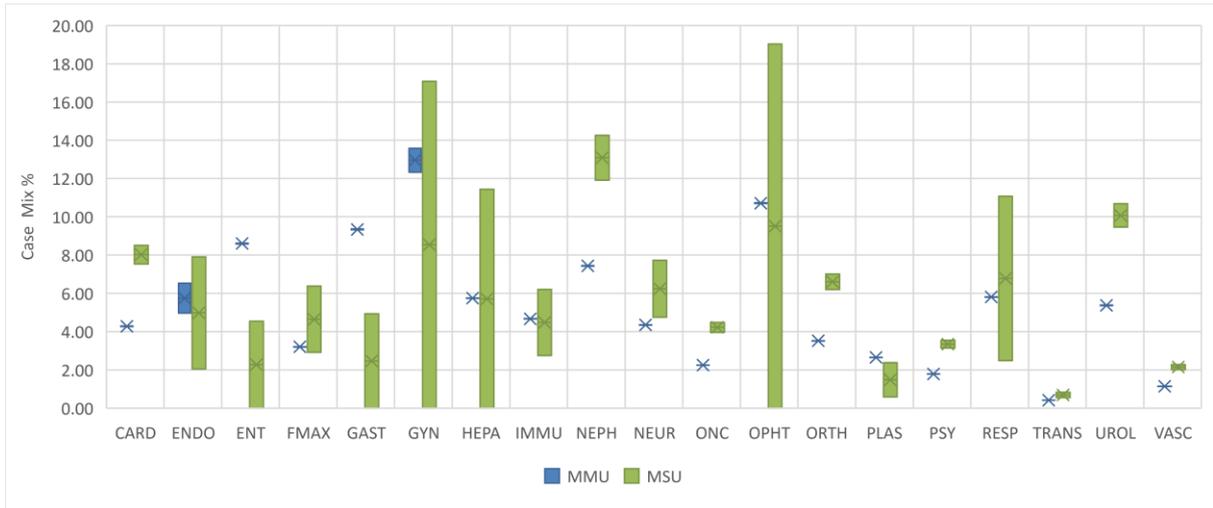

**Figure 27.** Differences observed in the case mix (UF13)

**UF14** ⬚. The results for UF14 are summarised in Figure 28 and 29. This function is most comparable to UF5 and UF13 and as such, the results are similar. For both MMU and MSU objectives the number of patients treated increases as the intercept is increased. The intercept forces higher outputs in some groups to be chosen. However, the total utility decreases quite significantly. This reduction occurs because more groups incur negative utilities as the intercept is increased.

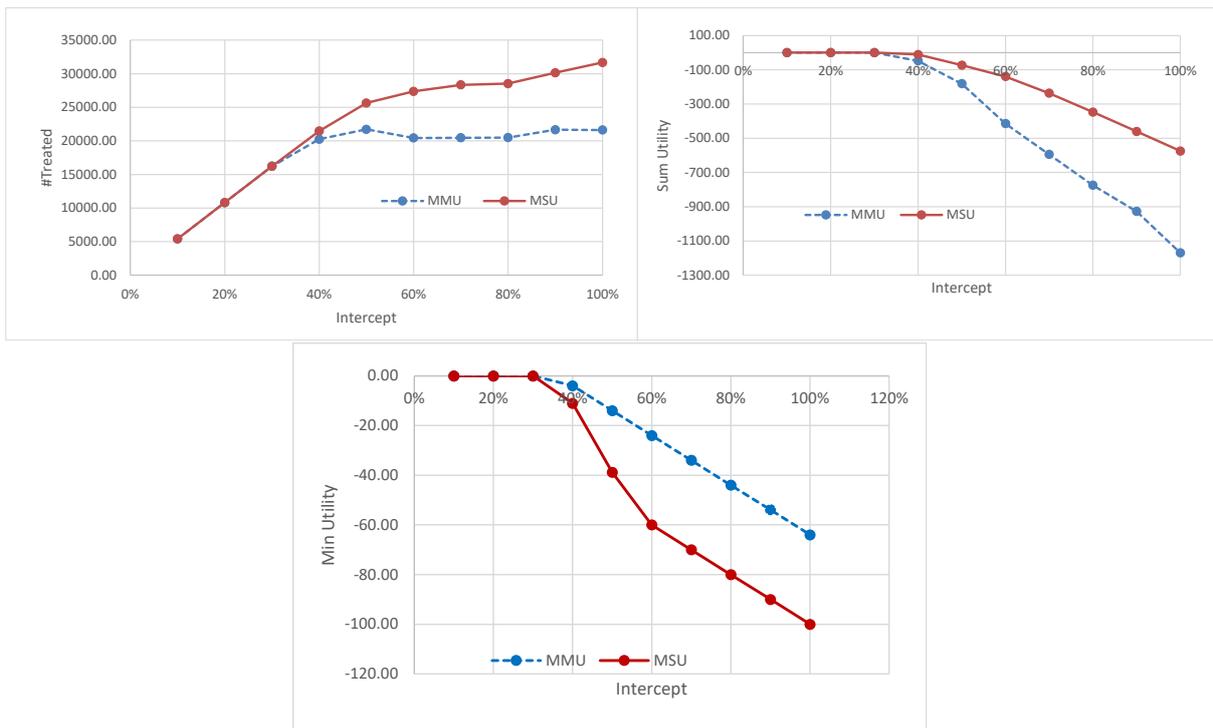

**Figure 28.** Intercept versus total treatments, sum of utility, and minimum utility for UF14

The case mix for the MMU objective is quite stable and only varies slightly as the intercept is increased. For the MSU objective, some specialties have increased presence in the caseload and some decrease in response. Only two are zeroed, however.

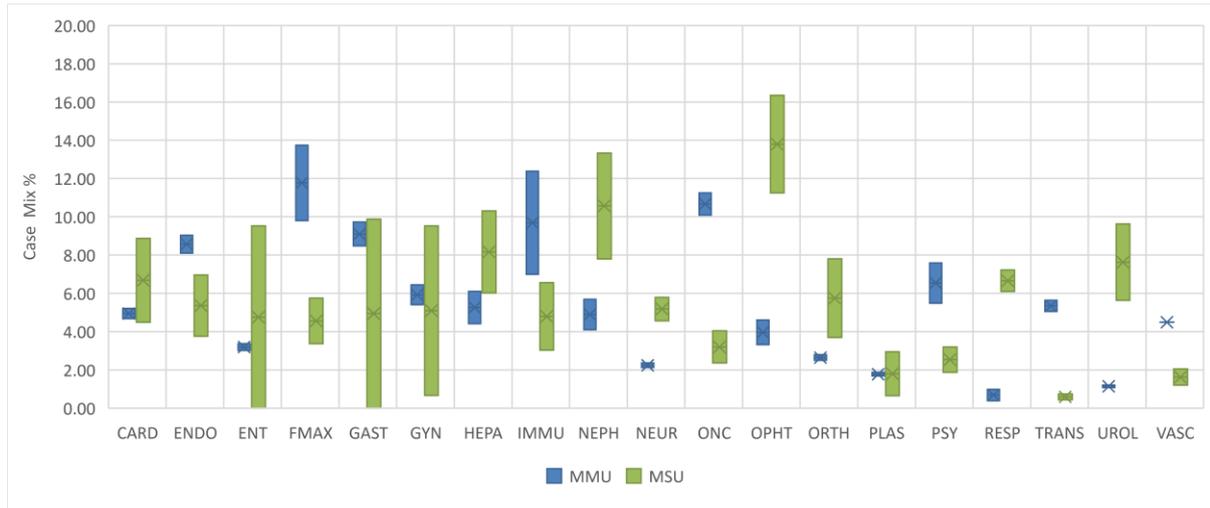

**Figure 29.** Differences observed in the case mix (UF14)

### 4.3. Pareto Optimality of Solutions

The Pareto optimality of caseloads obtained in section 4.2 were analysed. To check Pareto optimality the CMP was applied with the constraint $n_g^1 \geq \hat{n}_g^1 \ \forall g \in G$ and a Maximize $\mathbb{N}$ objective. The goals $\hat{n}_g^1$ were set as the existing caseload. The results are shown in Table 6. To note; i) the exact meaning of UF parameters has been explained previously, ii) a difference of zero means the caseload was Pareto optimal, and a positive difference means the caseload was dominated, iii) "zeroed caseload" means $n_g^1 = 0 \ \forall g \in G$ and a minimum group utility greater than zero could not be obtained. All zeroed caseloads are clearly not Pareto optimal.

**Table 6.** Pareto optimality of caseloads

| Method | $\mathbb{N}$ | | Diff | Diff (%) |
|---|---|---|---|---|
| | Initial | Corrected | | |
| GAM | 22389.66 | 33530.74 | 11141.07 | 49.76 |
| GPM (minimize max under) | 21232.76 | 32196.79 | 10964.03 | 51.64 |
| GPM (minimize sum under) | 31664.00 | same | 0 | 0 |

| Funct. | Param. | $\mathbb{N}$ (MMU) | | Diff. | $\mathbb{N}$ (MSU) | | Diff. |
|---|---|---|---|---|---|---|---|
| | | Initial | Corrected | | Initial | Corrected | |
| UF1 | $\alpha$=1 | 21610.42 | 30928.55 | 9318.13 | 31664.00 | same | 0 |
| | $\alpha$=0.15 | 20315.59 | 32232.47 | 11916.88 | 32001.87 | same | 0 |
| | $\alpha$ = 3 | 20460.73 | 30964.59 | 10503.86 | 29220.08 | same | 0 |
| UF2 | (10%) | 21537.21 | 30854.87 | 9317.67 | 32024.11 | same | 0 |
| | (90%) | zeroed caseload obtained | | | 25565.77 | 28241.53 | 2675.76 |
| UF3 | (10%) | 5407.79 | 34600.94 | 29193.15 | 5407.79 | 34600.94 | 29193.15 |
| | (100%) | 20807.70 | 30928.60 | 10120.89 | 31663.97 | same | 0 |
| UF4 | (5%, 95%) | 20492.85 | 30968.19 | 10475.34 | 30900.76 | 30952.98 | 52.21 |
| | (20%, 80%) | 19482.28 | 32442.66 | 12960.38 | 28521.94 | 28578.56 | 56.62 |
| | (40%, 60%) | zeroed caseload obtained | | | 28816.7 | 32418.74 | 3602.03 |
| UF5 | (10%) | 20372.80 | 31301.57 | 10928.77 | 31663.97 | same | 0 |
| | (100%) | 20551.92 | 30674.33 | 10122.41 | 31663.97 | same | 0 |
| UF6 | (40%) | 21245.39 | 31358.17 | 10112.78 | 21464.89 | 31517.39 | 10052.50 |
| | (100%) | 20734.49 | 30854.87 | 10120.38 | 31511.33 | 31522.99 | 11.66 |
| UF7 | (20, 0.1) | 21074.59 | 32232.47 | 11157.88 | 30249.15 | same | 0 |
| | (20, 0.9) | 19924.63 | 31978.08 | 12053.45 | 29220.06 | same | 0 |
| UF8 | (30%) | 16223.37 | 33357.72 | 17134.35 | 20977.59 | 32697.59 | 11720.00 |
| | (70%) | zeroed caseload obtained | | | 27081.41 | same | 0 |
| | (100%) | zeroed caseload obtained | | | 26908.58 | 26918.73 | 10.16 |
| UF11 | (10%) | 19482.28 | 32442.66 | 12960.38 | 32191.15 | same | 0 |
| | (100%) | 20263.33 | 30736.8 | 10473.47 | 31663.97 | same | 0 |
| UF12 | (10%) | 20263.33 | 30736.8 | 10473.47 | 31765.95 | same | 0 |
| | (60%) | zeroed caseload obtained | | | 29850.05 | 30984.91 | 1134.87 |
| | (100%) | zeroed caseload obtained | | | 30221.02 | same | 0 |
| UF13 | (10%) | 20413.87 | 31342.97 | 10929.10 | 32011.30 | same | 0 |
|  | (100%) | zeroed caseload obtained |  |  | 31645.24 | same | 0 |
| UF14 | (40%) | 20245.97 | 31860.15 | 11614.19 | 21464.89 | 31517.39 | 10052.50 |
|  | (100%) | 21610.42 | 30928.61 | 9318.19 | 31663.97 | same | 0 |

In summary, none of the caseloads obtained for the MMU objective were Pareto optimal. These caseloads, however, can be significantly improved as shown. Many caseloads for the MSU objective are Pareto optimal. Those not Pareto optimal were analysed and found to occur in specific conditions. Any patient type that lies on a flat segment of their associated UF results in the production of a dominated caseload. Examples include UF2, UF3, UF4, UF8, UF12, UF14. The other special case is UF6, which has the same utility value, for two different outputs.

### 4.4. Discussion and Insights

The numerical testing has provided the following insights.

(i). Once a UF "template" is chosen, it is recommended to perform a sensitivity to identify any insights that may be discernible. In some situations, UF parameters can have a greater effect on possible outputs. This sensitivity can be done even before specific parameters are chosen, and even when a particular parameter is of interest to a decision maker. For instance, there are insights regarding the aspiration level of individual specialties. The sensitivity analysis of UF3 and UF8 clearly shows that outputs below 36% of specialty capacity can easily be achieved. For higher aspiration levels the exact nature of the trade-offs that needs to be made (i.e., linear, or non-linear) can be observed. Aspiration and indifference levels may or may not affect total number of treatments. UF1, UF4, UF5, UF7, UF11, and UF12 are examples where total number of treatments can remain static. In other situations, those outputs can increase or decrease significantly. Some of the sensitivity analysis describe when regret for lost achievement becomes significant. For instance, regarding UF5, the regret manifested as negative utility, increases more greatly once aspirations exceed 70% of specialty capacity.

(ii). Those UF with static segments with utility zero (i.e., UF2, UF4, UF8, UF12) are nuanced in the sense that zeroed solutions may be identified for the MMU objective as optimal. Static segments (by definition), imply no significant difference over a range of values. If the level of indifference is too high, then many solutions will be regarded as having no utility. The CMP model will take advantage of this and choose the lowest output. To avoid zeroed solutions static segments should instead be sloped. This is not a problem for the MSU objective, however, as at least one group will have a non-zero caseload.

(iii). As the point of indifference is increased, there is a gradual decrease in output. The opposite occurs regarding the aspiration level, i.e., increased output occurs as it is increased.

iv). There are alternative optimal caseload solutions for some of the UF. For instance, the same $\sum_g u_g$ value can in theory be obtained in different ways, and this will lead to caseload solutions with different values of $\mathbb{N}$. The implication of this is that perhaps a tri-objective, with a $\epsilon_3 \mathbb{N}$ term could be considered worthwhile.

v). The MMU objective may be used as a primer to initiate the iterative process of multi-criteria CMP. The caseload produced is a good trade-off that gives minimal outputs to each group. In following stages of planning, reductions in those base levels may be traded, to allow other groups to prosper more. The MSU option automatically chooses some groups to prosper at the expense of others without negotiation.

vi). The convexity of the UF does not affect Pareto optimality of the caseload under the MSU objective. Only UF with flat segments and two values per utility are potentially dominated.

### 5. Conclusions

Most decision problems of any practical relevance involve the analysis of several conflicting criteria ([Makowski 2006](#)). Hospital case-mix planning is one such problem. In that task, it is necessary to apportion resources like operating theatres, wards, and intensive care units to different patient types, so that the maximum number of patients of each type is treated. There are, however, many ways to operate and that poses a significant dilemma to hospital managers and executives.

Current approaches for CMP involve the definition of a case mix and the search for a caseload, that abides by the proportions held within that case mix. Using some case mix concept, however, forces patient type outputs to be inherently interlinked. Dominated caseloads may occur because bottlenecks restricting the output of one group of patients, also cause reduced outputs of other patient types, even though free resources are available and additional outputs can be achieved.

Given perceived limitations in existing methods, utility functions (UF) were identified as a potential mechanism to facilitate improved hospital case-mix planning. This is a novel idea yet to be considered in the CMP literature, and a concept we believe managers and executives would find appealing. UF may be viewed as a means of regulating hospital capacity within a competitive environment, whereby the overall agenda is to treat as many patients of each type as possible. It is impossible to meet the needs of all patient types equally; hence treating a smaller number of patients is ultimately necessary. This is acceptable if individual patient types are treated in sufficient numbers. The application of UF to hospital case-mix planning has numerous other benefits. Utility functions can be used to model objective (i.e., quantitative) information including financial details, and subjective (i.e., qualitative) information like aspirations as well. Instead of treating non-achievement statically, as a linear relationship of the deviation from an aspiration, UF can model the varying importance of not meeting aspirations, which is more realistic. Other non-linear and discontinuous relationships can also be modelled using UF. In this article we considered linear/non-linear, monotonic/non-monotonic, convex/concave, simple/compound, and continuous/discontinuous utility functions.

The usage of utility functions is also intended as a means of avoiding the computationally intractable task of Pareto Front generation (i.e., identification of a set of non-dominated solutions) for a decision problem likely to have a high dimensional objective space. It is believed that utility functions and the application of a single objective involving minimum utility and aggregate utility is sufficient to provide a means of generating non-dominated solutions and navigating the space of optimal case mix options. It is worth noting that non-dominated solutions can be identified one by one, by altering the weights in equation (21).

Numerical testing indicates that the proposed approach has significant merit. Apart from finding desirable caseloads to consider, the approach can provide important insights via a sensitivity analysis of the parameters of each UF. The sensitivity analysis can provide a reality check to the managers and decision makers, as to the likely consequences of the choices that they are being asked to make.

The challenge of implementing an approach based upon utility functions concerns the creation of them. Inconsistent responses to questions regarding the nature of the UF being elicited may greatly affect the type of caseload determined. The omission of criteria and the confounding of criteria are two other drawbacks mentioned in the literature ([Stewart, 1996).](#) However, proper implementation and training may eliminate these issues. The definition of UF is synonymous to bidding in a competitive process. The chosen UF gives decision makers what they ask for. Having low aspirations early, may result in the CMP model overlooking the importance of specific patient groups, and permit prioritization of others more greatly.

**Future Research**: Basic utility functions of different types have been imposed and tested in this paper. Each specialty may, however, impose a completely different UF. It made no sense to make up scenarios of that nature as there are limitless possibilities. Pragmatically, however, it would be beneficial to test instances where the UF type is different. It would also be beneficial to run through the process of changing and negotiating UF for each specialty present in a hospital. This would help in the creation of a set of rigorous guidelines and processes, to optimise what is currently an ad-hoc process.

The UF considered in this article predominantly describe a linear relationship between output and utility, and as such they contain linear segments. A non-linear segment was, however, included in UF1. The non-linearity was shown to have significant effect on the resulting caseload. It would be beneficial to analyse the effect of non-linearity in all the UF discussed in this article.

This research facilitates the development of a decision support tool for hospital planners and executives. It would be worth exploring how such a tool could be designed and how users could interact with it.

CMP is typically performed with respect to a single criterion measured for each group of patients, i.e., the number of patients treated. It is possible to perform CMP when there is more than one kpi, i.e., there is a "vectorial return". Alternative kpi like revenue and cost could also be incorporated and used.

**Acknowledgements:** This research was funded by the Australian Research Council (ARC) Linkage Grant LP 180100542 and supported by the Princess Alexandra Hospital and the Queensland Children's Hospital in Brisbane, Australia.

## Appendix A. Acronyms

| Acronym | Meaning | Acronym | Meaning |
|---|---|---|---|
| **ARM** | Aspiration-Reservation Method | GAM, GPM | Goal Attainment and Goal Programming |
| **ASF** | Achievement Satisfaction/Scalarizing Function | LP | Linear Programming |
| **ASPT** | Aspiration Point | MCA | Multicriteria Analysis |
| **CMP** | Case-mix Planning | MCO | Multicriteria Optimization |
| **CUP** | Convex Up | MMU | Maximize Minimum Utility |
| **CDN** | Convex Down | MSU | Maximize Sum of Utility |
| **DM** | Decision Maker | PF | Pareto Frontier |
| **DRG** | Diagnosis Related Group | PTOI | Point of Indifference |
| **ECM** | Epsilon Constraint Method | UF and UFM | Utility Function and Utility Function Method |

## Appendix B. Details of piecewise linear utility functions

| Type | Description | Function | Ref Pt | Breakpoints and Gradients |
|---|---|---|---|---|
| UF1 | Linear Increasing - Basic | $u_g(n_g) = 100 \left(\frac{n_g}{\bar{n}_g}\right)$ for $n_g \leq \bar{n}_g$ | $(\bar{n}_g, 100.0)$ | $b_g = \{\bar{n}_g\}; \nabla_g = \{\frac{100.0}{\bar{n}_g}, 0.0\}$ |
| UF2 | Linear Increasing - Indifference | $u_g(n_g) = 100\left(\frac{\max(n_g - n_g^I, 0)}{\bar{n}_g - n_g^I}\right)$ | $(\bar{n}_g, 100.0)$ | $b_g = \{n_g^I\}; \nabla_g = \{0.0, \frac{100.0}{\bar{n}_g - n_g^I}\}$ |
| UF3 | Linear Increasing - Plateau | $u_g(n_g) = 100\left(\frac{\min(n_g, n_g^A)}{n_g^A}\right)$ | $(\bar{n}_g, 100.0)$ | $b_g = \{n_g^A\}; \nabla_g = \{\frac{100.0}{n_g^A}, 0.0\}$ |
| UF4 | Linear Increasing - Indifference - Plateau | $u_g(n_g) = \begin{cases} 100\left(\frac{n_g}{n_g^A - n_g^I}\right) & n_g^I \leq n_g \leq n_g^A \\ 100 & n_g > n_g^A \\ 0 & n_g < n_g^I \end{cases}$ | $(\bar{n}_g, 100.0)$ | $b_g = \{n_g^I, n_g^A\}; \nabla_g = \{0, \frac{100.0}{n_g^A - n_g^I}, 0.0\}$ |
| UF5 | Linear Increasing - Negative start | $u_g(n_g) = 100\left(\frac{n_g - n_g^I}{\bar{n}_g - n_g^I}\right)$ for $0 \leq n_g \leq \bar{n}_g$ | $(\bar{n}_g, 100.0)$ | $b_g = \{\bar{n}_g\}; \nabla_g = \{0, \frac{100.0}{\bar{n}_g - n_g^I}, 0.0\}$ |
| UF6 | Linear - Triangular | $u_g(n_g) = \begin{cases} 100\left(\frac{n_g}{n_g^A}\right) & 0 \leq n_g \leq n_g^A \\ 100\left(\frac{\bar{n}_g - n_g}{\bar{n}_g - n_g^A}\right) & n_g^A \leq n_g \leq \bar{n}_g \end{cases}$ | $(\bar{n}_g, 0.0)$ | $b_g = \{n_g^A\}; \nabla_g = \{\frac{100.0}{n_g^A}, -\frac{100.0}{\bar{n}_g - n_g^A}\}$ |
| UF8 | Discontinuous - One Tier | $u_g(n_g) = \begin{cases} 0 & n_g < n_g^I \\ 100 & n_g^I \leq n_g \leq \bar{n}_g \end{cases}$ | $(\bar{n}_g, 100.0)$ | $b_g = \{n_g^I, n_g^I\}; \nabla_g = \{0.0, 100.0, 0.0\}$ |
| UF9 | Discontinuous - Two Tier | $u_g(n_g) = \begin{cases} 0 & n_g < n_g^I \\ u_g^* & n_g^I \leq n_g \leq n_g^A \\ 100 & n_g^A \leq n_g \leq \bar{n}_g \end{cases}$ | $(\bar{n}_g, 100.0)$ | $b_g = \{n_g^I, n_g^I, n_g^A, n_g^A\};$ $\nabla_g = \{0.0, u_g^*, 0.0, 100 - u_g^*, 0.0\}$ |
| UF10 | Discontinuous - One Payoff | $u_g(n_g) = \begin{cases} 0 & n_g \neq n_g^A \\ 100 & n_g = n_g^A \end{cases}$ | $(0.0, 0.0)$ | $b_g = \{n_g^A, n_g^A, \bar{n}_g\};$ $\nabla_g = \{0.0, 100.0, -100.0, 0.0\}$ |
| UF11 | Linear - Regret | $u_g(n_g) = \begin{cases} n_g^1 f_g & n_g^1 \geq \hat{n}_g^1 \\ n_g^1 f_g - (\hat{n}_g^1 - n_g^1)\gamma_g & n_g^1 < \hat{n}_g^1 \end{cases}$ | $(\bar{n}_g, \bar{n}_g f_g)$ | $b_g = \{\hat{n}_g^1\}; \nabla_g = \{2f_g, f_g\}$ |
| UF12 | Discontinuous - Indifference - Jump | $u_g(n_g) = \begin{cases} w_g n_g^1 & n_g^1 \geq \hat{n}_g^1 \\ 0 & n_g^1 < \hat{n}_g^1 \end{cases}$ | $(\bar{n}_g^1, \bar{n}_g^1 w_g)$ | $b_g = \{\hat{n}_g^1, \hat{n}_g^1\}; \nabla_g = \{0.0, \hat{n}_g^1 w_g, w_g\}$ |
| UF13 | Discontinuous - Negative start -Jump | $u_g(n_g) = \begin{cases} w_g n_g^1 & n_g^1 \geq \hat{n}_g^1 \\ -\gamma_g(\hat{n}_g^1 - n_g^1) & n_g^1 < \hat{n}_g^1 \end{cases}$ | $(\bar{n}_g^1, w_g \bar{n}_g^1)$ | $b_g = \{\hat{n}_g^1, \hat{n}_g^1\}; \nabla_g = \{\gamma_g, w_g \hat{n}_g^1, w_g\}$ |
| UF14 | Compound - Negative start | $u_g(n_g) = \begin{cases} 0 & n_g^1 \geq \hat{n}_g^1 \\ -(\hat{n}_g^1 - n_g^1)\gamma_g & n_g^1 < \hat{n}_g^1 \end{cases}$ | $(\bar{n}_g, 0.0)$ | $b_g = \{\hat{n}_g^1\}; \nabla_g = \{\gamma_g, 0.0\}$ |

## Appendix C. Details of non-linear utility functions

| Type | Description | Function |
|---|---|---|

| UF1 | Non-linear variants | • $u_g(n_g) = 100e^{\alpha(n_g/\bar{n}_g^1)} - 1$ where $\alpha = \ln(101)$. Proof: $u_g(0) = e^{\alpha(0)} - 1 = 0$; $u_g(\bar{n}_g^1) = 100 \Rightarrow e^\alpha - 1 = 100 \Rightarrow e^\alpha = 101 \Rightarrow \alpha = \ln(101)$. The function produces only one curve with a convex shape. |
|---|---|---|

• $u_g(n_g^1) = 100(n_g^1/\bar{n}_g^1)^\alpha$. The curve is alterable; linear when $\alpha = 1$, convex up when $\alpha > 1$, convex down when $0 < \alpha < 1$, and constant when $\alpha = 0$.

• $u_g(n_g) = 100\left[1 - (1 - n_g/\bar{n}_g)^\beta\right]$. Linear when $\beta = 1$ convex up when $0 < \beta < 1$ and convex down when $\beta > 1$.

• $u_g(n_g) = \delta\left(e^{(n_g/\bar{n}_g)^\alpha} - 1\right)$ where $\delta = \frac{100}{e^1 - 1}$. The exponential term magnifies utility as $n_g \to \bar{n}_g$. The term $\delta$ calibrates the utility to 100 when $n_g = \bar{n}_g$. Function is concave when $\alpha$ is small (i.e., < 0.6), and convex when greater than two.

| UF2 | PTOI | $u_g(n_g) = 100\left(\frac{\max(n_g - n_g^I, 0)}{\bar{n}_g - n_g^I}\right)^\alpha$. Function is CUP or CDN depending on $\alpha$. |
|---|---|---|
| UF3 | ASPT | $u_g(n_g) = 100\left(\frac{\min(n_g, n_g^A)}{n_g^A}\right)^\alpha$. Function is CUP or CDN depending on $\alpha$. |
| UF6 | Triangular | $u_g(n_g) = \delta\left(\frac{n_g}{\bar{n}_g}\right)^\alpha \left(1 - \frac{n_g}{\bar{n}_g}\right)^\beta$. Function produces either convex or concave slopes, and asymmetry. Function needs further calibration to restrict range to [0,100]. |
| UF7 | S Shaped | $u_g(n_g) = \frac{100}{1 + e^{-\alpha \lambda_g}}$ where $\lambda_g = \left(\frac{n_g - n_g^R}{\bar{n}_g}\right) \equiv \frac{n_g}{\bar{n}_g} - \zeta_g$, $\alpha \in [1,100]$ and $\zeta_g \in [0,1]$. Parameter $\alpha$ describes how rapidly the utility changes, between 0 and 100. The reference point $n_g^R = \zeta_g \bar{n}_g$, indicates which value of $n_g^1$ has a utility of 0.5. If $n_g > n_g^R$ then $u_g = \frac{100}{1 + e^{-ve}}$ which implies the denominator is smaller and the utility is larger. If $n_g < n_g^R$ then $u_g = \frac{100}{1 + e^{+ve}}$ which implies the denominator is bigger and utility is smaller. When $\alpha \leq 10$ the domain is not [0,100]. For $\alpha \gg 10$ the function is appropriate as is. |

For the non-linear functions described it is necessary to compute breakpoints and gradients for each line segment in the following way:

$$b_g[i] = \Delta \times i \text{ where } \Delta = \frac{\bar{n}_g}{I-1} \text{ and } i = 1, \ldots, I-1$$

$$\nabla_g = \frac{(u_g(b_g[i]) - u_g(b_g[i-1]))}{\Delta} \text{ for } i = 1, \ldots, I$$

Regarding UF1, some parameter choices result in a UF2 or UF3 curve. However, these functions are not calibrated to specific $n_g^I$ and $n_g^A$ values.

**Appendix D.** Results of numerical testing – parameters of interest

| FUNCT. | PARAMETERS | MMU | | | MSU | | |
|---|---|---|---|---|---|---|---|
| | | $\mathbb{N}$ | $\sum_g u_g$ | $\min_g u_g$ | $\mathbb{N}$ | $\sum_g u_g$ | $\min_g u_g$ |
| UF1 | na | 21610.42 | 730.41 | 36.03 | 31663.97 | 1325.00 | 0.00 |
| | (ALPHA=2) | 20460.73 | 262.97 | 13.01 | 29220.06 | 1265.81 | 0.00 |
| | (ALPHA=3) | 20460.73 | 98.99 | 4.71 | 29220.06 | 1248.32 | 0.00 |
| | (ALPHA=0.15) | 20315.59 | 1660.22 | 85.79 | 32001.84 | 1769.74 | 82.43 |
| | (ALPHA=0.3) | 20979.04 | 1462.04 | 73.60 | 32088.76 | 1656.55 | 65.28 |
| UF2 | (PTOI@10%, ALPHA=1) | 21537.21 | 598.72 | 28.92 | 32024.11 | 1293.45 | 0.00 |
| | (PTOI@20%, ALPHA=1) | 20381.79 | 401.83 | 20.03 | 29160.40 | 1275.28 | 0.00 |
| | (PTOI@30%, ALPHA=1) | 19482.28 | 163.57 | 8.61 | 27912.98 | 1268.29 | 0.00 |
| | (PTOI@40%, ALPHA=1) | 0.00 | 0.00 | 0.00 | 28124.77 | 1263.00 | 0.00 |
| | (PTOI@50%, ALPHA=1) | 0.00 | 0.00 | 0.00 | 29220.06 | 1255.60 | 0.00 |
| | (PTOI@60%, ALPHA=1) | 0.00 | 0.00 | 0.00 | 31188.44 | 1244.51 | 0.00 |
| | (PTOI@70%, ALPHA=1) | 0.00 | 0.00 | 0.00 | 31142.47 | 1226.01 | 0.00 |
| | (PTOI@80%, ALPHA=1) | 0.00 | 0.00 | 0.00 | 25565.77 | 1200.00 | 0.00 |
| | (PTOI@90%, ALPHA=1) | 0.00 | 0.00 | 0.00 | 25565.77 | 1200.00 | 0.00 |
| UF3 | (ASPT@10%, ALPHA=1) | 5407.79 | 1900.00 | 100.00 | 5407.79 | 1900.00 | 100.00 |
| | (ASPT@20%, ALPHA=1) | 10815.58 | 1900.00 | 100.00 | 10815.58 | 1900.00 | 100.00 |
| | (ASPT@30%, ALPHA=1) | 16223.37 | 1900.00 | 100.00 | 19177.37 | 1900.00 | 100.00 |
| | (ASPT@40%, ALPHA=1) | 21245.40 | 1849.58 | 90.07 | 21092.44 | 1872.43 | 72.43 |
| | (ASPT@50%, ALPHA=1) | 24105.68 | 1674.62 | 72.05 | 24647.04 | 1752.94 | 22.43 |
| | (ASPT@60%, ALPHA=1) | 20438.27 | 1209.07 | 60.04 | 26573.89 | 1667.00 | 0.00 |
| | (ASPT@70%, ALPHA=1) | 26196.79 | 1282.89 | 51.47 | 27739.51 | 1561.70 | 0.00 |
| | (ASPT@80%, ALPHA=1) | 23495.59 | 990.22 | 45.03 | 28521.94 | 1465.84 | 0.00 |
| | (ASPT@90%, ALPHA=1) | 23028.57 | 860.08 | 40.03 | 30137.56 | 1388.14 | 0.00 |
| UF4 | @(5%, 95%) | 20492.85 | 682.70 | 34.47 | 30900.76 | 1335.74 | 0.00 |
| | @(10%, 90%) | 20845.55 | 659.74 | 32.53 | 30137.56 | 1349.16 | 0.00 |
| | @(15%, 85%) | 20845.55 | 618.27 | 30.04 | 29355.68 | 1366.12 | 0.00 |
| | @(20%, 80%) | 19482.28 | 507.50 | 26.71 | 28521.94 | 1387.78 | 0.00 |
| | @(25%, 75%) | 19482.28 | 419.00 | 22.05 | 27688.20 | 1418.11 | 0.00 |
| | @(30%, 70%) | 19482.28 | 286.25 | 15.07 | 26763.15 | 1464.52 | 0.00 |
| | @(35%, 65%) | 19482.28 | 65.00 | 3.42 | 27842.60 | 1513.90 | 0.00 |
| | @(40%, 60%) | 0.00 | 0.00 | 0.00 | 28816.70 | 1579.71 | 0.00 |
| UF5 | (INTERCEPT@5%) | 20381.79 | 638.38 | 32.66 | 31663.97 | 1294.74 | -5.26 |
| | (INTERCEPT@10%) | 20372.80 | 569.82 | 28.92 | 31663.97 | 1261.11 | -11.11 |
| | (INTERCEPT@20%) | 20372.80 | 403.54 | 20.03 | 31663.97 | 1181.25 | -25.00 |
| | (INTERCEPT@30%) | 20833.08 | 202.76 | 8.61 | 31663.97 | 1078.58 | -42.86 |
| | (INTERCEPT@40%) | 19482.28 | -125.83 | -6.62 | 31663.97 | 941.67 | -66.67 |
| | (INTERCEPT@50%) | 20734.49 | -480.35 | -27.95 | 31663.97 | 750.01 | -100.00 |
| | (INTERCEPT@60%) | 20381.79 | -1096.34 | -59.93 | 31663.97 | 462.51 | -150.00 |
| | (INTERCEPT@70%) | 20455.00 | -2089.90 | -113.25 | 31663.97 | -16.66 | -233.33 |
| | (INTERCEPT@80%) | 20381.79 | -4092.67 | -219.87 | 31663.97 | -974.99 | -400.00 |
| | (INTERCEPT@90%) | 20734.49 | -10001.77 | -539.74 | 31663.97 | -3849.97 | -900.00 |
| UF6 | (ASPT@10%) | 5407.79 | 1900.00 | 100.00 | 5407.79 | 1900.00 | 100.00 |
| | (ASPT@20%) | 10815.58 | 1900.00 | 100.00 | 10815.58 | 1900.00 | 100.00 |
| | (ASPT@30%) | 16223.37 | 1900.00 | 100.00 | 16223.37 | 1900.00 | 100.00 |
| | (ASPT@40%) | 21245.40 | 1849.58 | 90.07 | 21464.89 | 1872.43 | 72.43 |
| | (ASPT@50%) | 23588.94 | 1618.72 | 72.05 | 24647.04 | 1752.94 | 22.43 |
| | (ASPT@60%) | 20438.27 | 1209.07 | 60.04 | 25602.78 | 1667.00 | 0.00 |
| | (ASPT@70%) | 26196.79 | 1282.89 | 51.47 | 28554.27 | 1561.70 | 0.00 |
| | (ASPT@80%) | 23495.59 | 990.22 | 45.03 | 28521.94 | 1465.84 | 0.00 |
| | (ASPT@90%) | 23028.57 | 860.08 | 40.03 | 30137.56 | 1388.14 | 0.00 |
| UF7 | ($\lambda$=20, REF@10%) | 21074.59 | 1890.77 | 99.42 | 30249.15 | 1897.31 | 99.26 |
| | ($\lambda$=20, REF@20%) | 21179.69 | 1834.44 | 95.90 | 30249.15 | 1880.68 | 94.77 |
| | ($\lambda$=20, REF@30%) | 21083.60 | 1584.66 | 76.39 | 30249.15 | 1782.13 | 71.02 |
| | ($\lambda$=20, REF@40%) | 19870.72 | 633.44 | 31.58 | 30458.63 | 1666.23 | 0.03 |
| | ($\lambda$=20, REF@50%) | 19766.09 | 124.11 | 6.05 | 30458.63 | 1525.00 | 0.01 |
| | ($\lambda$=20, REF@60%) | 20061.20 | 20.16 | 0.90 | 28397.17 | 1359.54 | 0.03 |
| | ($\lambda$=20, REF@70%) | 20381.79 | 10.23 | 0.37 | 28825.19 | 1294.42 | 0.25 |
| | ($\lambda$=20, REF@80%) | 19924.63 | 34.52 | 1.81 | 28193.03 | 1270.58 | 1.80 |
| | ($\lambda$=20, REF@90%) | 19924.63 | 226.53 | 11.92 | 29220.06 | 1291.82 | 11.92 |

| FUNCT. | PARAMETERS | MMU | | | MSU | | |
|---|---|---:|---:|---:|---:|---:|---:|
| | | $\mathbb{N}$ | $\sum_g u_g$ | $\min_g u_g$ | $\mathbb{N}$ | $\sum_g u_g$ | $\min_g u_g$ |
| UF7 | ($\lambda$=30, REF@10%) | 20338.86 | 1899.24 | 99.95 | 30922.83 | 1899.82 | 99.94 |
| | ($\lambda$=30, REF@20%) | 23184.72 | 1886.81 | 99.09 | 30434.53 | 1896.41 | 98.72 |
| | ($\lambda$=30, REF@30%) | 20280.80 | 1641.38 | 84.79 | 30111.79 | 1837.43 | 79.33 |
| | ($\lambda$=30, REF@40%) | 19482.28 | 465.71 | 24.51 | 30250.80 | 1690.13 | 0.00 |
| | ($\lambda$=30, REF@50%) | 19482.28 | 32.06 | 1.69 | 30349.28 | 1575.04 | 0.00 |
| | ($\lambda$=30, REF@60%) | 19482.28 | 1.64 | 0.09 | 27568.37 | 1384.22 | 0.00 |
| | ($\lambda$=30, REF@70%) | 20021.86 | 0.73 | 0.02 | 28115.41 | 1298.55 | 0.01 |
| | ($\lambda$=30, REF@80%) | 14175.42 | 4.73 | 0.25 | 30842.08 | 1274.90 | 0.25 |
| | ($\lambda$=30, REF@90%) | 1266.77 | 90.11 | 4.74 | 29160.40 | 1236.01 | 4.74 |
| UF8 | (ASPT@10%) | 5407.79 | 1900.00 | 100.00 | 5407.79 | 1900.00 | 100.00 |
| | (ASPT@20%) | 10815.58 | 1900.00 | 100.00 | 10815.58 | 1900.00 | 100.00 |
| | (ASPT@30%) | 16223.37 | 1900.00 | 100.00 | 20977.59 | 1900.00 | 100.00 |
| | (ASPT@40%) | 0.00 | 0.00 | 0.00 | 21833.52 | 1800.00 | 0.00 |
| | (ASPT@50%) | 0.00 | 0.00 | 0.00 | 24833.81 | 1700.00 | 0.00 |
| | (ASPT@60%) | 0.00 | 0.00 | 0.00 | 28647.02 | 1500.00 | 0.00 |
| | (ASPT@70%) | 0.00 | 0.00 | 0.00 | 27081.41 | 1400.00 | 0.00 |
| | (ASPT@80%) | 0.00 | 0.00 | 0.00 | 24141.65 | 1300.00 | 0.00 |
| | (ASPT@90%) | 0.00 | 0.00 | 0.00 | 29004.62 | 1200.00 | 0.00 |
| UF11 | (ASPT@10%) | 19482.28 | 684.50 | 36.03 | 32011.30 | 1318.78 | 10.00 |
| | (ASPT@20%) | 19482.28 | 684.50 | 36.03 | 32191.15 | 1307.09 | 20.00 |
| | (ASPT@30%) | 19482.28 | 684.50 | 36.03 | 32224.61 | 1294.49 | 30.00 |
| | (ASPT@40%) | 21124.21 | 712.32 | 32.05 | 32325.07 | 1267.15 | 3.92 |
| | (ASPT@50%) | 19482.28 | 419.00 | 22.05 | 31906.23 | 1213.93 | -47.13 |
| | (ASPT@60%) | 20474.37 | 266.99 | 12.05 | 30921.67 | 1145.17 | -60.00 |
| | (ASPT@70%) | 19482.28 | 39.00 | 2.05 | 31110.89 | 1060.71 | -70.00 |
| | (ASPT@80%) | 19482.28 | -151.00 | -7.95 | 31577.78 | 959.43 | -80.00 |
| | (ASPT@90%) | 19482.28 | -341.00 | -17.95 | 31769.31 | 855.14 | -90.00 |
| UF12 | (PTOI@10%) | 20263.33 | 710.01 | 36.03 | 31765.95 | 1324.82 | 0.00 |
| | (PTOI@20%) | 20492.85 | 709.43 | 36.03 | 31849.36 | 1324.05 | 0.00 |
| | (PTOI@30%) | 20492.85 | 709.43 | 36.03 | 31803.66 | 1320.86 | 0.00 |
| | (PTOI@40%) | 0.00 | 0.00 | 0.00 | 32019.76 | 1317.32 | 0.00 |
| | (PTOI@50%) | 0.00 | 0.00 | 0.00 | 31358.47 | 1305.27 | 0.00 |
| | (PTOI@60%) | 0.00 | 0.00 | 0.00 | 29850.05 | 1282.87 | 0.00 |
| | (PTOI@70%) | 0.00 | 0.00 | 0.00 | 28036.70 | 1277.80 | 0.00 |
| | (PTOI@80%) | 0.00 | 0.00 | 0.00 | 29357.35 | 1276.38 | 0.00 |
| | (PTOI@90%) | 0.00 | 0.00 | 0.00 | 31338.53 | 1200.00 | 0.00 |
| **UF13** | (PTOI@10%) | 20413.87 | 702.73 | 36.03 | 32011.30 | 1318.78 | 10.00 |
| | (PTOI@20%) | 20492.85 | 709.43 | 36.03 | 32191.15 | 1307.09 | 20.00 |
| | (PTOI@30%) | 20492.85 | 709.43 | 36.03 | 32224.61 | 1294.49 | 30.00 |
| | (PTOI@40%) | 0.00 | 0.00 | 0.00 | 32019.76 | 1237.32 | -40.00 |
| | (PTOI@50%) | 0.00 | 0.00 | 0.00 | 31650.44 | 1204.69 | -50.00 |
| | (PTOI@60%) | 0.00 | 0.00 | 0.00 | 31176.37 | 1055.76 | -60.00 |
| | (PTOI@70%) | 0.00 | 0.00 | 0.00 | 29917.23 | 921.10 | -70.00 |
| | (PTOI@80%) | 0.00 | 0.00 | 0.00 | 28523.10 | 828.30 | -80.00 |
| | (PTOI@90%) | 0.00 | 0.00 | 0.00 | 31583.65 | 691.09 | -90.00 |
| **UF14** | (PTOI@10%) | 5407.79 | 0.00 | 0.00 | 5407.79 | 0.00 | 0.00 |
| | (PTOI@20%) | 10815.58 | 0.00 | 0.00 | 10815.58 | 0.00 | 0.00 |
| | (PTOI@30%) | 16223.37 | 0.00 | 0.00 | 16223.37 | 0.00 | 0.00 |
| | (PTOI@40%) | 20245.97 | -47.68 | -3.97 | 21464.89 | -11.03 | -11.03 |
| | (PTOI@50%) | 21712.08 | -181.66 | -13.97 | 25646.71 | -73.53 | -38.79 |
| | (PTOI@60%) | 20438.27 | -414.56 | -23.97 | 27373.63 | -139.80 | -60.00 |
| | (PTOI@70%) | 20461.83 | -594.56 | -33.97 | 28339.31 | -236.81 | -70.00 |
| | (PTOI@80%) | 20485.40 | -774.56 | -43.97 | 28521.94 | -347.33 | -80.00 |
| | (PTOI@90%) | 21664.38 | -927.18 | -53.97 | 30137.56 | -460.67 | -90.00 |